\documentclass{article}

\PassOptionsToPackage{numbers,sort&compress}{natbib}
\usepackage[final]{neurips_2022}

\usepackage[utf8]{inputenc} 
\usepackage[T1]{fontenc}    
\usepackage[colorlinks=true,linkcolor=black,citecolor=blue,urlcolor=blue,]{hyperref}     
\usepackage{url}            
\usepackage{booktabs}       
\usepackage{amsfonts}       
\usepackage{nicefrac}       
\usepackage{microtype}      
\usepackage{xcolor}         
\usepackage{bbding}
\usepackage{booktabs}
\usepackage{tablefootnote}
\usepackage[flushleft]{threeparttable}
\usepackage{graphicx}
\usepackage{caption}
\usepackage{subcaption}
\usepackage{mathrsfs}
\usepackage{makecell}
\usepackage{etoolbox}
\usepackage{multirow}
\usepackage{stfloats}
\usepackage{amssymb}
\usepackage{amsmath}
\usepackage{mathtools}
\usepackage{array}
\usepackage{algorithm}
\usepackage{algorithmic}
\usepackage{setspace}
\usepackage{hyperref}

\usepackage{bm}
\newcommand*{\B}[1]{\ifmmode\bm{#1}\else\textbf{#1}\fi}

\makeatletter
\def\thanks#1{\protected@xdef\@thanks{\@thanks
        \protect\footnotetext{#1}}}
\makeatother

\title{Learning Generalizable Models for Vehicle Routing Problems via Knowledge Distillation}
\author{
Jieyi Bi$^{1, \dagger}$\thanks{$^\dagger$ Equally contributed.}~,
Yining Ma$^{2, \dagger}$,
Jiahai Wang$^{1, \ast}$,
Zhiguang Cao$^{3, \ast}$\thanks{$^\ast$ Jiahai Wang and Zhiguang Cao are the corresponding authors.}~,
Jinbiao Chen$^{1}$,\\
~\textbf{Yuan Sun}$^4$,
and~\textbf{Yeow Meng Chee}$^2$\\
$^1$School of Computer Science and Engineering, Sun Yat-sen University\\
$^2$National University of Singapore\\
$^3$Singapore Institute of Manufacturing Technology, A*STAR\\
$^4$University of Melbourne\\
\texttt{bijy6@mail2.sysu.edu.cn},~ \texttt{yiningma@u.nus.edu}\\ 
\texttt{wangjiah@mail.sysu.edu.cn},~
\texttt{zhiguangcao@outlook.com}~\\
\texttt{chenjb69@mail2.sysu.edu.cn},
\texttt{yuan.sun@unimelb.edu.au},\\
\texttt{ymchee@nus.edu.sg}
}

\begin{document}

\maketitle

\begin{abstract}

Recent neural methods for vehicle routing problems always train and test the deep models on the same instance distribution (i.e., uniform). To tackle the consequent cross-distribution generalization concerns, we bring the \emph{knowledge distillation} to this field and propose an \emph{Adaptive Multi-Distribution Knowledge Distillation} (AMDKD) scheme for learning more generalizable deep models. Particularly, our AMDKD leverages various knowledge from multiple teachers trained on exemplar distributions to yield a light-weight yet generalist student model. Meanwhile, we equip AMDKD with an adaptive strategy that allows the student to concentrate on difficult distributions, so as to absorb hard-to-master knowledge more effectively. Extensive experimental results show that, compared with the baseline neural methods, our AMDKD is able to achieve competitive results on both unseen in-distribution and out-of-distribution instances, which are either randomly synthesized or adopted from benchmark datasets (i.e., TSPLIB and CVRPLIB). Notably, our AMDKD is generic, and consumes less computational resources for inference. 

\end{abstract}

\section{Introduction}
\label{sec:intro}

The \emph{Vehicle Routing Problem} (VRP) is a class of NP-hard combinatorial optimization problems with a wide variety of practical applications, such as freight delivery~\cite{Duan2020EfficientlyST}, last-mile logistics~\cite{konstantakopoulos2020vehicle} and ride-hailing~\cite{Qin2020RideHailingOD}.
For decades, the problem has been studied intensively in computer science and operations research, with numerous exact and (approximate) heuristic algorithms proposed \cite{lkh3,gurobi,Croes1958AMF,Shaw97anew}. Although the heuristic algorithms are usually preferred in practice given their
relatively higher computational efficiency, they heavily rely on hand-crafted rules and domain knowledge,
which may still leave room for improvement. As a promising alternative, deep (reinforcement) learning could be used to automatically learn a heuristic (or policy) for VRPs in an \emph{end-to-end} fashion, which has aroused widespread attention in recent years \cite{vinyals2015pointer,nazari2018reinforcement,kool2018attention,joshi2019efficient,kwon2020pomo,Kim2021LearningCP,ma2021learning,NEURIPS2021_3d863b36,9901466}. Compared to the traditional ones, the learned heuristics based on deep models could further reduce computational costs while ensuring desirable solutions.

However, existing deep models suffer from inferior generalization with respect to distributions of node coordinates. To be concrete, they often train and test neural networks on instances of the same distribution, mostly the uniform distribution, where deep models are able to achieve competitive results more efficiently than the traditional heuristics (e.g., \cite{kwon2020pomo}). Nevertheless, when the learned policy is applied to infer the \textit{out-of-distribution} (OoD) instances, the solution quality is usually low. This cross-distribution generalization issue inevitably hinders the applications of deep models, especially because the real-world VRP instances may follow various and even unknown distributions.

A number of preliminary attempts have been made to tackle this generalization issue for VRPs, which mainly leverage (automatic) data augmentation \cite{Zhang2022LearningTS,xin2022generative,adversarial_iclr2022,wang2022game} and distributionally robust optimization~\cite{Jiang2022LearningTS}, respectively. However, they are not optimal in our view, as the former always starts with a specified single distribution which may limit the resulted performance, while the latter needs to manually define major and minor instances. Different from them, in this paper, we aim to enhance the cross-distribution generalization by transferring various policies learned from respective distributions into one, where we exploit \emph{knowledge distillation} to learn more generalizable models for VRPs.

Specifically, we propose a generic \emph{Adaptive Multi-Distribution Knowledge Distillation} (AMDKD) scheme for training a light-weight model with favorable cross-distribution generalization performance. To impart broad yet specialized knowledge, we exploit multiple teacher models that are (pre-)trained on respective \emph{exemplar} distributions. Our AMDKD then leverages those teachers to train a shared student model in turns, inspired by the learning paradigm of human-beings. Meanwhile, we also equip the AMDKD with an adaptive strategy to track the real-time learning performance of the student model on every exemplar distribution, which allows the student to concentrate on absorbing those hard-to-master knowledge, so as to strengthen the effectiveness of the learning.

Accordingly, our contributions are summarized as follows: (1) We bring the \emph{knowledge distillation} to the field of \emph{neural combinatorial optimization} and aim at improving the cross-distribution generalization for solving VRPs, which offers a promising perspective to transfer knowledge/policy learned from multiple models into one; (2) We propose the \emph{Adaptive Multi-Distribution Knowledge Distillation} (AMDKD) scheme, where we distill diverse knowledge from multiple teachers trained on exemplar distributions to yield a light-weight yet generalist student model, and also present an adaptive strategy for the student to better assimilate hard-to-master knowledge from difficult distributions; (3) We apply our generic AMDKD to two representative deep models, i.e., AM \cite{kool2018attention} and POMO \cite{kwon2020pomo}, respectively. Results show that, while consuming less computational resources for inference, our AMDKD performs favorably against the backbone models as well as other existing generalization methods for deep models, {especially on the benchmark dataset CVRPLIB~\cite{uchoa2017new}}. By further coupling with the efficient active search (EAS) \cite{eas}, our AMDKD achieves the new state-of-the-art performance. We also conduct a series of analysis to verify our designs.

\section{Related work}
\label{sec:relatedwork}
Recently, neural methods based on deep models for VRPs have aroused widespread interest. These methods are categorized as neural \emph{construction} models and neural \emph{improvement} models in general.

\textbf{Neural construction models.} 
They exploit deep (reinforcement) learning to autoregressively construct the solution in an end-to-end fashion. \citet{vinyals2015pointer} introduced the first RNN-based Pointer Network (Ptr-Net) to solve TSP based on supervised learning.
\citet{bello2016neural} then exploited reinforcement learning to train the Ptr-Net for TSP. 
In \cite{nazari2018reinforcement}, the Ptr-Net was then
extended to solve CVRP. Different from the above RNN-based methods, \citet{kool2018attention} proposed the well-known Attention Model (AM) based on the Transformer architecture \cite{vaswani2017attention}.
Subsequently, many studies extended AM for routing problems, such as \cite{peng2020corr,xin2020multi,li2021deep,kwon2020pomo,kwon2021matrix,Kim2021LearningCP,zhang2022meta}. As a representative, \citet{kwon2020pomo} proposed the POMO (Policy Optimization with Multiple Optima) and achieved significantly better performance. In a recent work, POMO was adapted to an efficient active search (EAS) framework to further boost the performance during inference~\cite{eas}.
Besides, graph neural networks (GNNs) are also utilized to effectively learn and identify the graph-structured features of the problem instance \cite{dai2017learning,joshi2019efficient,fu2020generalize,Duan2020EfficientlyST,kool2021deep}.

\textbf{Neural improvement models.}
They iteratively improve the initial solution by exploiting deep (reinforcement) learning to assist or control the local search. \citet{chen2019learning} proposed the NeuRewriter that learned a policy to partially rewrite the current solution. \citet{hottung2019neural,hottung2022neural} leveraged deep networks to learn to perform the large neighborhood search. \citet{wu2021learning} 
and \citet{d2020learning} 
proposed to control the 2-opt operation~\cite{Croes1958AMF}. \citet{ma2021learning} further upgraded the deep model of \citet{wu2021learning} to dual-aspect collaborative Transformer (DACT) with much superior performance. The DACT method was further enhanced in \cite{ma2022efficient} in order to learn a ruin-and-repair operation for pickup and delivery problems.
Generally, improvement methods consume much longer inference time than construction ones to achieve higher solution quality.

\textbf{Cross-distribution generalization.}
The above neural methods for VRPs often train and test the deep models on the same instance distribution. Although the state-of-the-art deep models perform close to or even surpass the strong traditional heuristics, they are incapable to generalize the learned policy to other distributions, which seriously impairs their practical applications. Therefore, the cross-distribution issue has gradually gained more attention~\cite{Zhang2022LearningTS,xin2022generative,adversarial_iclr2022,wang2022game,Jiang2022LearningTS}. Among the early attempts in this line of research, \citet{xin2022generative} proposed a generative adversarial network (GAN) based framework to generate hard-to-solve instances for training the model. \citet{wang2022game} leveraged the game theory and proposed the Policy Space Response Oracle (PSRO) framework to simultaneously learn a trainable solver and an instance generator. \citet{Zhang2022LearningTS} designed a hardness-adaptive TSP instance generator and adopted curriculum learning to train the model. These methods primarily emphasized on data augmentation or are limited to TSP. Different from them, \citet{Jiang2022LearningTS} adopted CNN to acquire distribution-aware features and exploited group DRO (Distributionally Robust Optimization) to enhance model robustness against distributions. However, this method needs to manually define major and minor instances. Note that we acknowledge that the generalization to different problem scales (or sizes) is also important, which we leave as the future research direction.

\section{Preliminaries and notations}
\label{sec:model}
We first present the formulation of the studied VRPs and classic distributions. Then we introduce how deep models are used to solve them, followed by the basic rationale of knowledge distillation.

\subsection{VRPs and their distributions}
\label{sec:problem}
We define VRPs over a complete graph $\mathcal{G}=\lbrace \mathcal{V}, \mathcal{E}\rbrace$, where $v_i \in \mathcal{V}$ represents the (customer) node, $e(v_i, v_j)\in\mathcal{E}$ represents the edge between two nodes, and $C[e(v_i, v_j)]$ represents the cost (we use \emph{length} in this paper) of the edge. By referring tour $\tau$ (a.k.a. solution) to a permutation of nodes in $\mathcal{V}$, the objective is usually to find the optimal tour $ \tau^*$ with the least total cost(length) over a finite search space $S$ containing all possible tours, which could be formulated as Eq.~(\ref{eq:opt}) in general,
\begin{equation}
    \tau^* = \mathop{\arg\min}_{\tau'\in\mathcal{S}} {L(\tau{'} \vert \mathcal{G})} = \mathop{\arg\min}_{\tau'\in\mathcal{S}} \sum_{e(v_i, v_j) \in \tau'} C\left[e(v_i, v_j)\right].
    \label{eq:opt}
\end{equation}
For different VRP variants, such objective may be subject to different problem-specific constraints~\cite{chen2022deep,zhang2021solving}, where multiple sub-tours may exist in a valid tour $\tau$. Following the recent literature~\cite{kool2018attention,kwon2020pomo,eas}, we focus on two representative VRPs, i.e., TSP and CVRP, respectively. A feasible tour for TSP considers a vehicle visiting each node in $\mathcal{V}$ once and only once. As an extension from a vehicle to a fleet, CVRP considers an extra depot node $v_0$ and a capacity limit $Q$ for any given vehicle, where each customer node $v_i (i=1,...,n)$ is associated with a demand request $\delta_i$.
A feasible tour for CVRP consists of multiple sub-tours, each of which represents a vehicle in the fleet departing from the depot, serving a subset of nodes, and finally returning to the depot. The total demand of each sub-tour must not exceed the capacity $Q$, and all nodes except for the depot must be visited once and only once.

\begin{figure}
    \centering
    \includegraphics[width=0.85\textwidth]{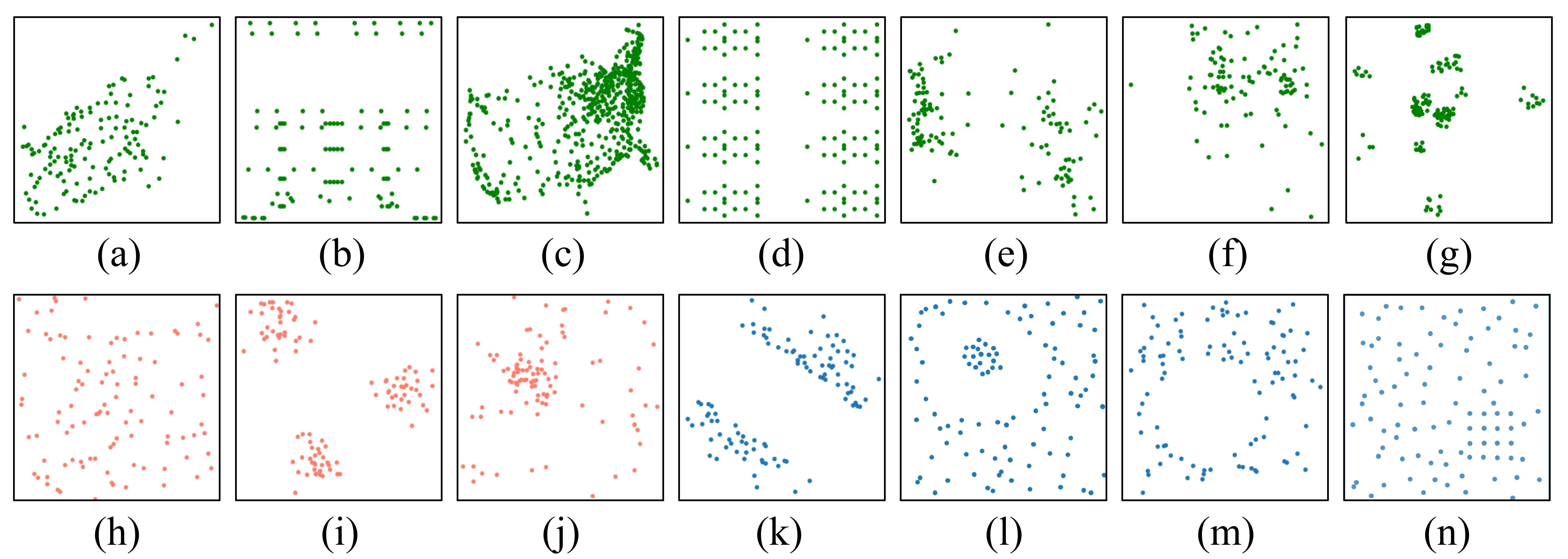}
    \caption{{VRP instances following various distributions from the literature: (a) gr137, (b) lin105, (c) att532, (d) pr136, (e) X-n125-k30, (f) bier127, (g) Tai150d, (h) Uniform, (i) Cluster, (j) Mixed, (k) Expansion, (l) Implosion, (m) Explosion, (n) Grid, where instances (a)-(g) are from TSPLIB~\cite{reinelt1991tsplib} and CVRPLIB~\cite{uchoa2017new}. In this paper, we consider instances following distributions (h)-(j) for training and other unseen distributions (k)-(n), as well as unseen benchmark datasets for testing.}}
    \label{fig:dist}
\end{figure}

In Figure~\ref{fig:dist}, we visualize a number of VRP instances following various distributions from the literature \cite{kool2018attention,Bossek2019EvolvingDT,Jiang2022LearningTS}, including the TSPLIB~\cite{reinelt1991tsplib} and CVRPLIB~\cite{uchoa2017new} benchmark datasets. As can be observed, the node coordinates of a VRP instance may follow complicated and even unknown distribution, which considerably intensifies the hardness for solving. While the recent neural methods report superior performance to the traditional heuristics on some fixed distributions, it is unfortunate that they are more sensitive to distribution shift. It is thus of great importance and interest to develop powerful deep models that can simultaneously handle as many diverse distributions as possible.

\subsection{Tour construction by deep models}
\label{sec:drl}
We focus on neural construction methods for VRPs, which usually exploit deep neural networks to sequentially construct the tour. In light of this, the solving procedure is mostly modelled as a Markov Decision Process (MDP), where the Transformer styled \cite{kool2018attention} architectures following the encoder-decoder structure are often adopted as the policy network. Typically, the encoders project the nodes of the instance into node embeddings for feature extraction. Afterwards, the decoder builds the tour $\tau$ based on learned node embeddings and the partial tour $a_{1:t-1}$ constructed previously. At time step $t$ of the MDP, the decoder picks an unvisited node as the action $a_t$ where invalid ones are masked for feasibility. The procedure is repeated until the whole tour is completed, which is factorized as,
\begin{equation}
    p_{\theta}(\tau \vert \mathcal{G}) = \prod\limits_{t=1}^{\ell} p_{\theta}(a_t \vert a_{1:t-1}, \mathcal{G}),
    \label{eq:cons}
\end{equation}
where $p_\theta$ is the policy parameterized by $\theta$, and $\ell$ denotes the final time step in MDP.
For TSP, $\ell=n$; for CVRP, $\ell\geq n$ since the depot can be visited more than once\footnote{The sub-tours for CVRP are constructed sequentially. At each step, the agent selects an unvisited node whose demand is smaller than the remaining capacity or returns to the depot for full replenishment.}. The total reward is defined as the negative of the tour length, i.e., $-L$($\tau \vert \mathcal{G}$). This is essentially consistent with the objective in Eq. (\ref{eq:opt}). 

\subsection{Knowledge distillation}
\label{sec:pre-kd}

Knowledge distillation \cite{hinton2015distilling} is a kind of teacher-student training paradigm that aims at transferring knowledge from a (group of) complex teacher model(s) $\theta^\text{T}$, to a succinct student model $\theta^\text{S}$. The recent research findings reveal that knowledge distillation is not only able to effectively learn a lighter student network from larger teacher network(s) \cite{jiao-etal-2020-tinybert,liu-etal-2020-fastbert}, but also has potential to improve the generalization  \cite{stanton2021does,NEURIPS2020_2288f691} even over its teacher(s) \cite{hu-etal-2018-attention,pmlr-v80-furlanello18a,Berseth2018ProgressiveRL}. Typically, the teacher models are pre-trained, and the student model compares the output of teacher model(s) with its own and considers it as a supervisory signal. Formally, the student network is trained with the goal of minimizing a weighted combination of the distillation loss $\mathcal{L}_\text{KD}$ and the original task loss $\mathcal{L}_\text{Task}$ as follows,
\begin{equation}
    \mathcal{L} = \alpha\mathcal{L}_\text{Task} + (1-\alpha)\mathcal{L}_\text{KD},
    \label{total_loss}
\end{equation} 
where $\alpha \in [0,1]$. The $\mathcal{L}_\text{KD}$ with respect to one or more teachers ($N_T\!\geq\!1$) is generally formulated as,
\begin{equation}
    \mathcal{L}_\text{KD} = \frac{1}{N_T}\sum\limits_{x\in\mathcal{X}}\sum\limits_{i=1}^{N_T}\phi\left[p_{\theta_{i}^\text{T}}(x), p_{\theta^\text{S}}(x)\right],
    \label{eq:multi}
\end{equation}
where $x$ is the training data from $\mathcal{X}$ and $\phi(\cdot)$ measures the statistical distance between the teacher and the student, such as the Kullback-Leibler divergence $\text{KL}( p_{\theta^\text{T}} \Vert  p_{\theta^\text{S}}) = \sum_i p_{\theta^\text{T}} (\log  p_{\theta^\text{T}} - \log  p_{\theta^\text{S}}$).

\section{Methodology}
\label{sec:methodology}
Consider what happens in a classroom when students study multiple modules. Typically, they (students) learn only one module (exemplar distribution) at a time from a professional teacher, and then go on to the next one until all modules are mastered. When students do poorly on a quiz after class (validation dataset), they will be required to study that module more frequently in order to become generalists. Motivated by this, we propose the \emph{Adaptive Multi-Distribution Knowledge Distillation} (AMDKD) scheme that could efficiently learn more generalizable models for VRPs.

Previously, deep models were trained on a single distribution, e.g., the Uniform in~\cite{kwon2020pomo,Kim2021LearningCP,ma2021learning}.
More recently, attempts have been made to expand the training data by augmenting atypical (minor)~\cite{Jiang2022LearningTS} or hard instances~\cite{xin2022generative,Zhang2022LearningTS}. Differently, AMDKD allows expert knowledge for tackling distinct distributions to be transferred into a single yet generalist model via distillation. With a light network and high inference speed, the resulted model is supposed to perform favorably against existing methods. Moreover, AMDKD is generic, and could be used to boost various deep models for VRPs.

\begin{figure}
    \centering
    \includegraphics[width=0.99\textwidth]{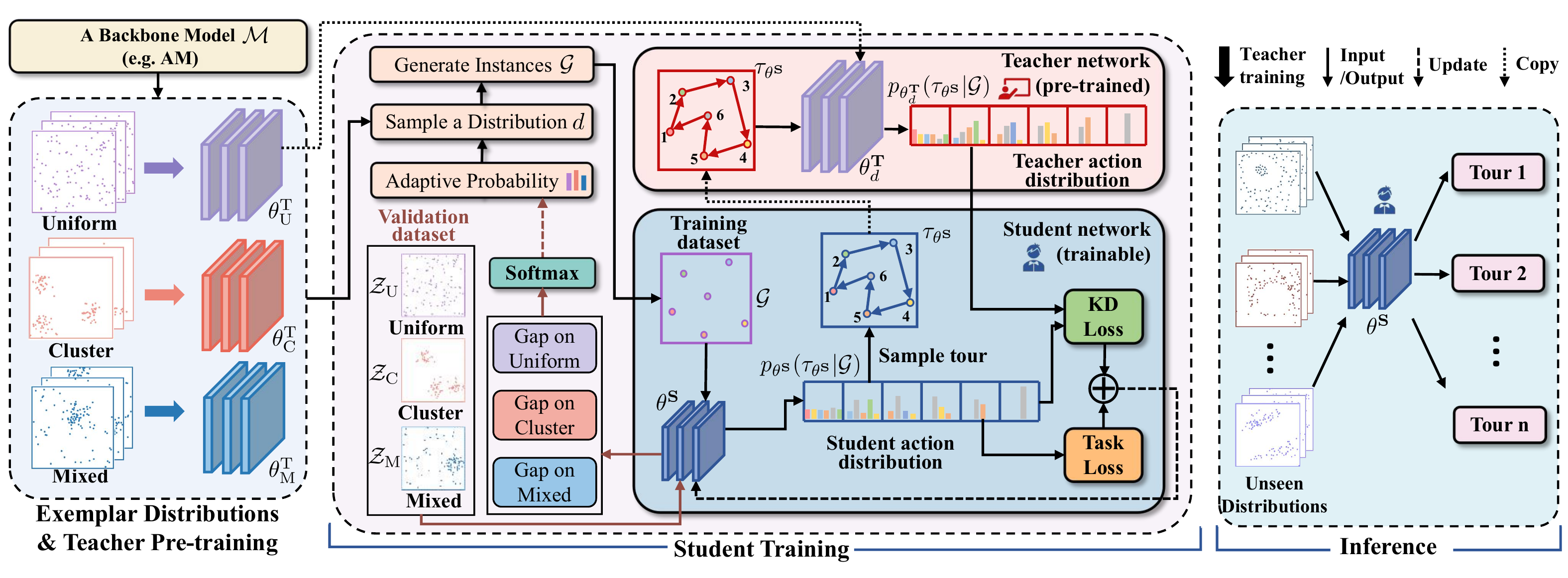}
    \caption{Framework of our AMDKD scheme. From left to right: teacher pre-training, student training and inference. The Uniform is selected as the current exemplar distribution for an example.}
    \label{fig:architecture}
\end{figure}

\subsection{Overall structure and teacher (pre)-training}
\label{sec:teacher_training}

The overview framework of our AMDKD is illustrated in Figure \ref{fig:architecture}. Given an existing deep model (e.g., AM \cite{kool2018attention}), AMDKD first performs teacher training (or directly use pre-trained ones) to obtain a teacher model for each distribution listed in a set of \emph{exemplar} ones. Then in each knowledge transfer epoch, AMDKD picks a specific distribution $d$ and its teacher $\theta^\text{T}_d$
to train a student network. To facilitate effective learning, the likelihood of picking each distribution is adaptively updated according to the current performance of the student on a validation dataset.
Finally, AMDKD performs on-policy distillation, allowing the student network $\theta^\text{S}$ to sample RL trajectories (i.e., tours $\{\tau_{\theta^\text{S}}\}$) for training, and calculates the two losses (i.e., KD loss and Task loss) for an update.

We consider Uniform, Cluster, and Mixed (mixture of uniform and cluster) as exemplar distributions to train teacher networks in this paper. Consequently, a set of well-trained teachers with parameters $\theta^\text{T} = \lbrace\theta^\text{T}_\text{U}, \theta^\text{T}_\text{C}, \theta^\text{T}_\text{M}\rbrace$ are attained for each distribution.
Despite only limited distributions are exploited throughout training, the learned student network is expected to absorb generic and robust knowledge from teachers that may effectively generalize to unseen distributions such as Expansion, Implosion, Explosion, and Grid (visualization in Figure~\ref{fig:dist}). 
Our motivation here is that each teacher only needs to master a specific distribution, and by working together via our distilling scheme, they can educate a generalist student. 
{Note that the exemplar distributions used here could also be substituted with others (see Section \ref{sec:ablation_md})}. Furthermore, since our designed distillation scheme is model-agnostic, the teacher network can follow most of existing architectures. In this paper, we assess our scheme by applying it to two representative construction methods, i.e., AM \cite{kool2018attention} and POMO \cite{kwon2020pomo}, respectively.

\subsection{Adaptive multi-distribution student training}
\label{sec:adaptive}

Given the well-trained teachers on exemplar distributions, our AMDKD student selects one teacher and one distribution in each knowledge transfer epoch and gradually learns to make appropriate decisions.
Note that there is another line of works that argue for the simultaneous use of multiple teachers in the distillation process~\cite{Rusu2016PolicyD,wu2021one}. In their strategy, all teachers are engaged to provide a weighted loss to train the student as aforementioned in Section \ref{sec:pre-kd}. However, such design may require teachers to be trained on a homogeneous task primarily with supervised learning, which does not suit our heterogeneous distributions with reinforcement learning {(see Appendix \ref{app:ablation} for results)}.

\textbf{Student network.} 
We stipulate that the student shares a similar architecture with its teachers, but can reduce its network parameters as needed to speed up the inference. Nevertheless, we find in Section \ref{sec:ablation_md} that a larger student model usually leads to a better performance. Therefore, there is a trade-off between solution quality and computational cost. In this paper, we consider reducing the dimension of the node embeddings from $128$ (teacher) to $64$ (student), resulting in a reduction of 61.8\% and 59.2\% in the model parameters for the adaption with AM and POMO, respectively. We also considered lowering the number of encoder layers, but the results were below the expectation.

\textbf{The adaptive multi-distribution distilling strategy.} 
In each epoch, AMDKD selects one distribution and its corresponding teacher model. At the beginning, it selects each distribution with an equal probability. After epoch $E'$ (a hyper-parameter), the probability would be adaptively adjusted according to the performance of the student on the given validation datasets $\mathcal{Z}_\text{U},\mathcal{Z}_\text{C},\mathcal{Z}_\text{M}$ (each with 1,000 instances) for each exemplar distribution. The likelihood $p^\text{adaptive}$
of selecting distribution $d\in \{U,C,M\}$ is proportional to the exponent value of the gaps to the LKH solver~\cite{lkh3} as follows, 
\begin{equation}
p^\text{adaptive}(d) = \left\{\begin{matrix}
\text{Softmax}    \left({\text{AvgGap}\left[\{\tau_{\theta^\text{S}}\}|_{\mathcal{Z}_d},\{\tau_{\text{solver}}\}|_{\mathcal{Z}_d}\right]}\right), & \text{if } E \geq E'\\
\frac{1}{|\mathcal{D}|}, & 
\text{otherwise}
\end{matrix}\right.
    \label{eq:softmax}
\end{equation}
where $|\mathcal{D}|$ is the number of exemplar distributions;
$\{\tau_{\theta^\text{S}}\}|_{\mathcal{Z}_d}$
 and $\{\tau_{\text{solver}}\}|_{\mathcal{Z}_d}$ refer to the tour set
 generated by the student $\theta^\text{S}$ and the LKH solver on dataset $\mathcal{Z}_d$, respectively.
The validation sets are fixed, hence the LKH solver only needs to run once to attain $\{\tau_{\text{solver}}\}|_{\mathcal{Z}_d}$ for $d\in\{U,C,M\}$.
{Meanwhile, we note that $p^\text{adaptive}$ may converge during distillation (see Appendix~\ref{app:ablation}), which means that it is possible to stop such evaluation early and reuse the stabilized one for an even faster training.}

\begin{algorithm}[t]
    \caption{Adaptive Multi-Distribution Knowledge Distillation (AMDKD)}
    \begin{algorithmic}[1]
        \REQUIRE {A backbone model $\mathcal{M}$ (e.g., AM), 
        exemplar distributions $\mathcal{D}$ (e.g., $\mathcal{D} = \{U, C, M\}$).
        }
        \STATE {Randomly initialize teacher networks $\theta^\text{T}_d$ ($\forall d \in \mathcal{D}$) and student network $\theta^\text{S}$ according to $\mathcal{M}$;}
        \STATE {Perform teacher training or leverage pre-train ones (if any) to attain well-learned $\theta^\text{T}_d$ ($\forall d \in \mathcal{D}$);}
        \FOR {$\text{epoch} = 1,2,...,E$}
            \STATE {Pick distribution $d$ and its teacher $\theta^\text{T}_d$ with an adaptive probability according to Eq. (\ref{eq:softmax});}
            \FOR {$\text{step} = 1,2,...,T$}
                \STATE {
                Let student $\theta^\text{S}$ sample  tours $\tau^i_{\theta^\text{S}}$ for each  $\{\mathcal{G}_i\}_{i=1}^B$ according to its own policy $p_{\theta^\text{S}}$;}
                \STATE {Get $\nabla \mathcal{L}_\text{Task}$ by estimating Eq.~(\ref{eq:taskloss}) as per the original design of $\mathcal{M}$;}
                \STATE {Get $\nabla \mathcal{L}_\text{KD}$ by computing the gradients of Eq.~(\ref{eq:KDloss});}
                \STATE {$\theta^\text{S} \gets \theta^\text{S} + \eta\nabla \mathcal{L}$} where $\nabla \mathcal{L} \gets \alpha\nabla \mathcal{L}_\text{Task} + (1-\alpha)\nabla \mathcal{L}_\text{KD} $.  
            \ENDFOR
            \ENDFOR
    \end{algorithmic}
        \label{algorithm}
\end{algorithm}

\textbf{Loss function.}
Following Eq. (\ref{total_loss}), we leverage the task loss ($\mathcal{L}_\text{Task}$) and the KD loss ($\mathcal{L}_\text{KD}$) to jointly train the student with $\alpha\!=\!0.5$. Pertaining to $\mathcal{L}_\text{Task}$, we define the task loss as follows,
\begin{equation}
 \mathcal{L}_\text{Task} = -J(\theta^\text{S}|d) = -\mathbb{E}_{\mathcal{G} \sim d, \tau \sim p_{\theta^\text{S}}(\tau \vert \mathcal{G})}[L(\tau|\mathcal{G})],
 \label{eq:taskloss}
\end{equation}
where the training instances $\mathcal{G}$ are sampled following the selected distribution $d$, and the tours $\tau$ are constructed via the student network $\theta^\text{S}$ according to Eq. (\ref{eq:cons}). In AM and POMO, the REINFORCE algorithm~\cite{Williams2004SimpleSG} is used to estimate the gradients for the above loss function and we follow exactly the same way as per their original designs.
Pertaining to $\mathcal{L}_\text{KD}$, it is defined to encourage the student to imitate how a teacher network sequentially selects the nodes. Specifically, given the instance $\mathcal{G}$ sampled from distribution $d$ and a tour $\tau_{\theta^\text{S}}$ constructed by the student $\theta^\text{S}$ following its own policy, our AMDKD leverages $p_{\theta^\text{T}_d}(\tau_{\theta^\text{S}} \vert \mathcal{G})$ suggested by the teacher network $\theta^\text{T}_d$, to compute $\mathcal{L}_\text{KD}$ that measures the similarity of the probability distributions between teacher and student using the KL divergence,
\begin{equation}
    \mathcal{L}_\text{KD} = \frac{1}{B}\sum_{i=1}^B\sum_{a_j \in \tau_{\theta^\text{S}}}p_{\theta_d^\text{T}}(a_j \vert \mathcal{G}_i) \left(\log   {p_{\theta_d
    ^\text{T}}(a_j \vert \mathcal{G}_i)} - \log{p_{\theta^\text{S}}(a_j \vert \mathcal{G}_i)} \right).
    \label{eq:KDloss}
\end{equation}

We summarize our AMDKD in Algorithm \ref{algorithm}. Note that our AMDKD follows the on-policy scheme where the tours $\tau_{\theta^\text{S}}$ are output by the student that currently performs learning. As an alternative, such tours can also be output by the teacher (i.e., off-policy scheme). However, it performs inferior to our on-policy one {(see Section \ref{sec:ablation_md})}.
Finally, we note that the student learned by our AMDKD could also be coupled with the \emph{efficient active search} (EAS)~\cite{eas} during inference to further boost the performance. Accordingly, the resulting AMDKD+EAS achieves a new state-of-the-art performance.

\section{Experiments}
\label{sec:exp}
We conduct experiments on TSP and CVRP with $n=$ 20, 50, and 100 nodes similar to~\cite{kwon2020pomo,ma2021learning}. As aforementioned, we adopt Uniform, Cluster and Mixed (mixture of uniform and cluster) as exemplar distributions for training; Expansion, Implosion, Explosion, and Grid as the unseen distributions for testing. For the above 7 distributions, we follow \cite{kool2018attention,Bossek2019EvolvingDT,Jiang2022LearningTS} to generate the respective instances (details are presented in Appendix \ref{app:dist}). All experiments are conducted on a machine with NVIDIA RTX 3090 GPU cards and Intel Xeon Silver 4216 CPU at 2.10GHz. {Our implementation in PyTorch are publicly available\footnote{\url{https://github.com/jieyibi/AMDKD}}. Some additional analysis and discussions can be found in Appendix~\ref{app:ablation}.}

\begin{table}[t]
  \centering
  \caption{Distillation effectiveness of AMDKD on three exemplar distributions.}\vspace{6pt}
  \resizebox{0.99\textwidth}{!}{
    \begin{threeparttable}
    \begin{tabular}{cl|c|cccc|cccc|cccc}
    \toprule
    \multicolumn{2}{c|}{\multirow{2}[2]{*}{Model}}   &  \multicolumn{1}{c|}{{Size}} & \multicolumn{4}{c|}{{$n$ = 20}} & \multicolumn{4}{c|}{{$n$ = 50}} & \multicolumn{4}{c}{{$n$ = 100}} \\
\multicolumn{2}{c|}{} & (M) & \multicolumn{1}{c}{$\text{G}_\text{U}$} & \multicolumn{1}{c}{$\text{G}_\text{C}$} & \multicolumn{1}{c}{$\text{G}_\text{M}$} & \multicolumn{1}{c|}{Avg.}& \multicolumn{1}{c}{$\text{G}_\text{U}$} & \multicolumn{1}{c}{$\text{G}_\text{C}$} & \multicolumn{1}{c}{$\text{G}_\text{M}$} & \multicolumn{1}{c|}{Avg.}& \multicolumn{1}{c}{$\text{G}_\text{U}$} & \multicolumn{1}{c}{$\text{G}_\text{C}$} & \multicolumn{1}{c}{$\text{G}_\text{M}$} & \multicolumn{1}{c}{Avg.} \\
    \midrule
        \multicolumn{1}{c}{\multirow{8}[8]{*}{\rotatebox{90}{{TSP}}}} 
     & AM(U) & 0.68 & 0.09\% & 0.26\% & 0.19\% & 0.18\% &  0.59\% & 2.24\% & 1.36\% & 1.39\% &  2.10\% & 7.49\% & 4.06\% & 4.55\%\\
            & AM(C) & 0.68 & 0.17\% & 0.10\% & 0.27\% & 0.18\% &  1.41\% & {0.80\%} & 2.14\% & 1.45\% &  3.76\% & 6.97\% & 4.39\% & 5.04\% \\
            & AM(M) & 0.68 & 0.15\% & 0.16\% & 0.13\% & 0.15\% &  1.19\% & 1.71\% &  {0.87\%} & 1.26\% &  3.08\% & {5.65\%} & {2.55\%} & 3.76\% \\\cmidrule{2-15}     
            & AMDKD-AM &  \textbf{0.26} &  {0.02\%}  &  {0.06\%}  & {0.05\%}  &  \textbf{0.04\%}  &   {0.25\% }    &   {1.64\% }   &  {0.86\%}  &   \textbf{0.91\%}  &  {1.21\% }    &   {5.63\% }   &  {3.55\%}  &   \textbf{3.46\%} \\

\cmidrule{2-15}          
& POMO(U) & 1.20 &0.00\% & 0.01\% & 0.01\% & 0.01\%&    {0.04\%} & 0.42\% & 0.21\% & 0.22\%&    0.17\% & 1.97\% & 0.92\% & 1.02\% \\
          & POMO(C) & 1.20 &0.00\% & 0.00\% & 0.01\% & 0.00\%&    0.09\% & 0.07\% & 0.21\% & 0.12\%&    0.41\% & {0.29\%} & 0.83\% & 0.51\%\\
          & POMO(M) & 1.20 &0.00\% & 0.01\% & 0.00\% & 0.00\% &  {0.08\%} & 0.17\% & {0.08\%} & 0.11\% &  0.77\% & 1.17\% & {0.34\%} & 0.76\% \\\cmidrule{2-15}     
          & AMDKD-POMO &  \textbf{0.49} &{0.00\%} & {0.00\%} & {0.00\%} & \textbf{0.00\%} &    0.05\% & {0.05\%} & {0.09\%} &  \textbf{0.06\%} &  
          {{0.34\%}} & {{0.35\%}} & {{0.41\%}} & \textbf{{{0.37\%}}}\\
    \midrule
 \multicolumn{1}{c}{\multirow{8}[8]{*}{\rotatebox{90}{{CVRP}}}} & AM(U)   & 0.68 & 1.98\%  & 1.99\%  & 1.98\%  & 1.98\%  & 2.53\%  & 4.33\%  & 2.99\%  & 3.28\%  & 3.10\%  & 9.87\%  & 4.57\%  & 5.85\%  \\
          & AM(C)& 0.68& 1.62\%  & 1.43\%  & 1.74\%  & 1.60\%  &    3.08\%  & 2.75\%  & 3.35\%  & 3.06\%  &     4.27\%  & {3.89\%}  & 4.93\%  & 4.36\%  \\
          & AM(M) & 0.68& 2.09\%  & 2.19\% & 2.05\%  & 2.11\%  & 2.74\%  & 3.17\%  & 2.31\%  & 2.74\%      & 3.95\%  & 6.26\%  & 3.41\%  & 4.54\%    \\\cmidrule{2-15}     
& AMDKD-AM & \textbf{0.26} & {0.53\%}  & {0.59\%}  & {0.64\%}  & \textbf{0.59\%}  &  {1.61\%}  & {2.66\%} & {1.92\%} & \textbf{2.07\%} & {2.08\%} & 5.06\%  & {3.01\%} & \textbf{3.38\%} \\

\cmidrule{2-15}          & POMO(U) & 1.20 &0.36\%  & 0.49\%  & 0.51\%  & 0.45\%  &     {0.80\%}  & 1.53\%  & 1.07\%  & 1.13\%  &     0.95\%  & 2.34\%  & 1.31\%  & 1.53\%        \\
          & POMO(C) & 1.20 &0.41\%  & 0.40\%  & 0.54\%  & 0.45\%       & 1.16\%  & {0.93\% } & 1.07\%  & 1.05\%  &  {0.93\%}  & {1.28\%}  & 1.21\%  & {1.14\%}  \\
          & POMO(M) & 1.20 &0.36\%  & 0.51\%  & {0.40\%}  & 0.42\%  &    1.22\%  & 1.34\%  & {0.85\%}  & {1.14}\%  &     1.89\%  & 2.07\%  & {0.96\%}  & 1.64\%    \\\cmidrule{2-15}     
			
& AMDKD-POMO &  \textbf{0.49} &{0.35\%}  & {0.40\%}  & 0.41\%  & \textbf{0.39\%}  &  0.81\% & 0.97\% & 0.89\% & \textbf{0.89\%} &      {1.06\% }	& {1.36\%} & {0.99\% }	& \textbf{1.13\%}     \\
    \bottomrule
    \end{tabular}
    \begin{tablenotes}
    \item {\textit{Note:} Unless otherwise stated, the gaps are computed w.r.t. the strong traditional solvers Gurobi~\cite{gurobi} (for TSP) and LKH~\cite{lkh3} (for CVRP).}
    \end{tablenotes}
    \end{threeparttable}
    }
  \label{tab2}%
\end{table}%

\textbf{Training and hyper-parameters.} 
We apply our AMDKD to two representative deep models, i.e., AM \cite{kool2018attention} and POMO \cite{kwon2020pomo}, termed as {AMDKD-AM} and {AMDKD-POMO}, respectively.
{For the teacher (pre)-training phase, we directly follow the RL algorithm, network architecture, and other hyper-parameters as suggested in the original backbone method. For the student distillation phase, we use batch size $B$ = 512\footnote{For training AMDKD-POMO on TSP100 and CVRP100, we use $B$ = 128 due to GPU memory constraint.}, and task-specific hyper-parameters $T$ = 250, $E'$ = 500 for AMDKD-AM and $T$ = 20, $E'$ = 1 for AMDKD-POMO, respectively.
By default, the dimension of the node embeddings in our student networks AMDKD-AM and AMDKD-POMO is reduced from 128 (teacher) to 64 (student) to enjoy a faster inference speed.} 
The total numbers of needed training epochs vary with the problem size. Regarding {AMDKD-AM}, we use 2,000, 5,000, 10,000 for problem sizes 20, 50, 100, respectively (both TSP and CVRP); regarding {AMDKD-POMO}, we use 5,000 (TSP-20), 10,000 (CVRP-20, TSP-50, TSP-100), and 30,000 (CVRP-50, CVRP-100). The Adam optimizer is used with learning rate 1e-4. Training time also varies with the problem size. Taking CVRP-100 as an example, one epoch takes about 4 minutes for AMDKD-AM and 1.4 minutes for AMDKD-POMO.

\textbf{Inference.} 
For AMDKD-AM, it samples 1,280 solutions following AM \cite{kool2018attention}; and for {AMDKD-POMO}, we adopt the greedy rollout with $\times$8 augments following POMO~\cite{kwon2020pomo}. All experiments are conducted on test datasets containing 10,000 instances per distribution. Unless otherwise stated, the gaps are computed w.r.t. the strong traditional solvers Gurobi~\cite{gurobi} (for TSP) and LKH~\cite{lkh3} (for CVRP).

\subsection{Effectiveness analysis of AMDKD}
We first study the effectiveness of our AMDKD when applied to backbone model AM and POMO for TSP and CVRP, respectively. In Table \ref{tab2}, we display the gaps of the learned student and their respective teachers on unseen instances following three exemplar distributions (denoted as $\text{G}_\text{U}$, $\text{G}_\text{C}$, and $\text{G}_\text{M}$) and also report the parameter sizes of each model. We can see that when the teacher model is trained on a specific distribution, it does not generalize well on the other ones (especially for AM), resulting in poor overall performance (if referring to the average gaps). In contrast, our AMDKD could alleviate this issue effectively. While the learned student model by AMDKD reduces the size of the teacher model from 0.68 to 0.26 M (a 61.8\% reduction) for AM and from 1.20 to 0.49 M (a 59.2\% reduction) for POMO, it still exhibits improved overall performance for both TSP and CVRP on all the three sizes. Pertaining to AMDKD-AM, our AMDKD student not only significantly outperforms its three teachers for both TSP and CVRP,
but also exhibits even lower gaps on the distribution where the teacher was previously trained in most cases.  Pertaining to AMDKD-POMO, although POMO itself presents a better generalization, our AMDKD can still boost its overall performance with a much lighter student network for both TSP and CVRP. The above results validate that our AMDKD does not overfit to a single distribution, and successfully learns a lightweight yet generalist student model under the guidance of multiple teachers with expertise in different exemplar distributions.

\subsection{Generalization analysis of AMDKD}
We now assess the cross-distribution generalization of our AMDKD. In addition to the
two backbone models, we also compare with the following methods, i.e., {{1) other deep models}} including- 
{1.a) DACT} \cite{ma2021learning},
a neural \emph{improvement} model that learns to perform local search;
{1.b) LCP}~\cite{Kim2021LearningCP} (TSP only),
a {hybrid} two-stage method that combines \emph{improvement} and \emph{construction} strategy; 
and {{2) other methods specialized for generalization}} including-
{
2.a) HAC}~\cite{Zhang2022LearningTS} (TSP only), 
a fine-tuning framework for AM by leveraging instances with different hardness;
{2.b) DROP}~\cite{Jiang2022LearningTS}, a \emph{distributionally robust optimization} based method to enhance POMO;
2.c) GANCO~\cite{xin2022generative}, a framework that enhances AM by learning a \emph{generative adversarial network} to produce hard-to-solve training instances; and 
2.d) PSRO (a.k.a. LIH)~\cite{wang2022game}, an \emph{improvement} method with a \emph{game theory} based policy space response oracle framework. More implementation details of them are provided in Appendix~\ref{app:baseline}.
Note that for DROP, GANCO, and PSRO, we only compare the results on several benchmark instances reported in their original papers (see Table~\ref{tab:libsummary}), since their codes (for re-training) are not publicly available.  

\begin{table}[t]
  \centering
  \caption{Generalization on unseen in-distribution (ID) and out-of-distribution (OoD) instances.}
  \vspace{6pt}
  \resizebox{0.99\textwidth}{!}{
  \begin{threeparttable}
    \begin{tabular}{cl|c|ccc|ccc|ccc}
    \toprule
        \multicolumn{2}{c|}{\multirow{2}[1]{*}{Model}} & Size & \multicolumn{3}{c|}{$n=$ 20} & \multicolumn{3}{c|}{$n=$ 50} & \multicolumn{3}{c}{$n=$ 100} \\
          & & (M) & $\text{G}_\text{ID}$  & $\text{G}_\text{OoD}$  & Time$^*$  &  $\text{G}_\text{ID}$  &  $\text{G}_\text{OoD}$  & Time$^*$ &  $\text{G}_\text{ID}$  & $\text{G}_\text{OoD}$  & Time$^*$ \\
    \midrule
    \multicolumn{1}{c}{\multirow{11}[11]{*}{\rotatebox{90}{{TSP}}}} & Gurobi & - &   -  &   - &   0.01s (7s) &  - &   - &  0.08s (51s)  &   -  &   - & 0.7s (7.6m)\\
    \cmidrule{2-12}
    & HAC$^\#$  & 0.68 &  0.11\%  &  0.07\%   & 0.03s (48s)   &  1.05\%  &  0.57\%  &  0.09s (4m)  & 4.68\%  & 2.97\% & 0.19s (16m) \\
    & LCP$^\#$   & 2.03  &  0.11\%    &   0.03\%    &  1.2s (43m)  &   6.70\%    &  0.99\%     &  1.5s (2.8h)  &   32.70\%    &   8.24\%  & 2.4s (6.4h)  \\
    & DACT(T=1,280)$^\#$ & 0.27 &   0.11\%   &   0.08\%    &   12s (2.1m)   &   0.31\%      &  0.34\%     &   19s (7.2m)    &   2.37\%    &   3.13\%  & 28s (23m)  \\
    \cmidrule{2-12}
    & AM$^\#$ {(128)}    & 0.68 &    0.18\%   &   0.10\%    &  0.03s (48s)     &  1.48\%  &   0.74\%   & 0.09s (4m)   & 4.15\%  &  2.84\%  & 0.19s (16m)\\
    & {AM$^\#$ (64)} & 0.26 &   0.21\%   &   0.11\%    &  0.03s (35s) & {1.74\%}  & {0.83\%}  &   0.08s (2.8m)    &  {5.93\%} & {3.85\%} & 0.17s (11m)\\
    & {AMDKD-AM (64)} & 0.26 &   \textbf{0.04\%}   &   \textbf{0.02\%}    &  0.03s (35s) & \textbf{0.91\%}  & \textbf{0.37\%}  &   0.08s (2.8m)    &  \textbf{3.46\%} & \textbf{1.87\%} & 0.17s (11m)\\
    \cmidrule{2-12}
    & POMO$^\#$ {(128)}   & 1.20 &  0.00\%     &   0.00\%    &  0.03s (5s)  &  0.07\%     &  0.05\%  &   0.09s (16s)   &    {\textbf{0.30\%}}  & {\textbf{0.28\%}}  & 0.13s (1.1m)  \\
    & {POMO$^\#$ (64)} & 0.49 &    0.02\%   &   0.01\%     &  0.02s (4s)   &   0.18\%   &    0.16\%    & 0.04s (11s) & {0.69\%}  & 0.59\%      &  0.12s (50s) \\
    & {AMDKD-POMO (64)} & 0.49 &    \textbf{0.00\%}   &   \textbf{0.00\%}     &  0.02s (4s)   &   \textbf{0.06\%}   &    \textbf{0.05\%}    & 0.04s (11s) & {0.37\%}  & {0.41\%}      &  0.12s (50s) \\
    & {AMDKD+EAS$^\dagger$}  & 0.49 &  \textbf{0.00\%}  & \textbf{0.00\%}   &   5s (4.5m)    &  \textbf{0.01\%} &  \textbf{0.01\%}  & 12s (28m) &  \textbf{0.11\%}   & \textbf{0.10\%}  &  28s (2.3h) \\
    \midrule
    \multicolumn{1}{c}{\multirow{9}[9]{*}{\rotatebox{90}{{CVRP}}}} 
    & LKH3  &   -    &  - & - & 7.7s (1.3h) &   -   &  -    &   31s (5.3h) & - &  -  & 56s (9.6h)\\
    \cmidrule{2-12}
    & DACT(T=1,280)$^\#$ & 0.27 &   {0.09\%}    &    {0.05\%}    &  28s (4.3m)  &   {1.59\% }   &   {1.59\%}    &   55s (14m)    &   {5.56\%}    &  {5.55\%}  & 1.5m (34m) \\
    \cmidrule{2-12}
    & AM$^\#$ {(128)}   & 0.68 &    2.00\%  &   1.98\%  & 0.05s (1.2m)  &   3.39\%  &  {2.66\%} &  0.12s (5m)     &  5.42\%  &  3.75\% &  0.26s (18m)   \\
    &  {AM$^\#$ (64)}   & 0.26 & 2.02\%    &   2.02\%    & 0.05s (49s)    & 3.62\% &  2.65\%   & 0.11s (3.6m)  &    6.83\%     &  4.56\%    & 0.23s (13m) \\
    & {AMDKD-AM (64)} & 0.26 & \textbf{0.59\%}    &   \textbf{0.55\%}    & 0.05s (49s)    & \textbf{2.07\%} &  \textbf{1.69\%}   & 0.11s (3.6m)  &    \textbf{3.38\%}     &  \textbf{2.44\%}    & 0.23s (13m) \\
    \cmidrule{2-12}
    & POMO$^\#$ {(128)}  & 1.20 &  0.42\% &  0.39\%   & 0.05s (7.8s)      & 0.92\% &   0.94\%   &  0.10s (18s)  & 1.14\% &   1.21\%  & 0.19s (1.3m)\\
    & {POMO$^\#$ (64)} & 0.49 &  0.55\%  &  0.51\%     &   0.05s (6.1s)    & 1.19\% &  1.21\% & 0.08s (15s)  &  1.43\% &  1.50\%  & 0.18s (1.1m)\\
    & {AMDKD-POMO (64)} & 0.49 &  \textbf{0.39\%}  &  \textbf{0.36\%}     &   0.05s (6.1s)    &      \textbf{0.89\%} &  \textbf{0.90\%} & 0.08s (15s)  &  \textbf{1.13\%} &  \textbf{1.21\%}  & 0.18s (1.1m)\\
    & {AMDKD+EAS$^\dagger$}   & 0.49 &  \textbf{-0.04\%} & \textbf{-0.06\%} &  9s (7.8m) &  \textbf{0.07\%} &  \textbf{0.04\%}  &  20s (37m)     &  \textbf{-0.03\%} &  \textbf{-0.04\%} &  40s (3.3h)   \\       
    \bottomrule
    \end{tabular}%
    \begin{tablenotes}
\item[$*$] We report the average time to solve one instance, and the total time to solve 10,000 instances in ($\cdot$) with batch parallelism allowed (one GPU).
\item[$\#$] The corresponding model is trained on a mixed training dataset that contains instances from all the three exemplar distributions.
\item[$\dagger$] For EAS, we adopt its EAS-lay version (T=100) for demonstration purpose.
\end{tablenotes}
\end{threeparttable}
    }
    \label{tab1}
\end{table}%
\textbf{Distribution mixture augmentation.} Given that baselines AM, POMO, DACT, and LCP were originally trained on the uniform distribution only (resulting in poor generalization), we re-train the above models (with $^\#$) on a mixed dataset containing instances from the three exemplar distributions, to ensure a fair comparison. We also apply this mixed dataset as the initial training instances for HAC (with $^\#$). {Though such distribution mixture augmentation improves the generalization of construction methods POMO and AM on larger instances, we notice that it does not always happen so, especially for the methods that have an improvement component (e.g., DACT and LCP). We present detailed results and discussions regarding this finding in Appendix~\ref{app:baseline}. Meanwhile, for AM$^\#$ and POMO$^\#$, we report their performance with network dimensions of both 128 and 64 (same size as our AMDKD student) for a more comprehensive comparison.} We assess the following metrics, 1) parameter size; 2) average gap on unseen in-distribution instances, i.e., $\text{G}_\text{ID}$; 3) average gap on unseen out-of-distribution instances, i.e., $\text{G}_\text{OoD}$; 4) average time for solving an individual instance and total time for solving 10,000 instances with batch parallelism allowed (one GPU)\footnote{Even so, we note that the time of traditional solvers and deep models might be still difficult to be fairly compared due to different implementations (C vs Python) and computing devices (CPU vs GPU).}.

As can be observed in Table \ref{tab1},  our AMDKD is able to achieve competitive results on both unseen
in-distribution and out-of-distribution instances, and performs favourably against all baselines. Pertaining to TSP, our AMDKD-AM (64) significantly outperforms the baseline AM$^\#$ (128) and AM$^\#$~(64) on all sizes in terms of both $\text{G}_\text{ID}$ and $\text{G}_\text{OoD}$ with fewer parameters and higher inference speed. Meanwhile, it also consistently beats the recent baseline HAC$^\#$ (designed to boost the generalization of AM) on all sizes in terms of both two gaps and the time efficiency. {For the backbone model POMO, which suffers from less generalization concern, the baseline POMO$^\#$ (128) can already attain near-optimal solutions to TSP instances with a desirable generalization performance. However, our AMDKD (64) is able to achieve
{competitive}
generalization results with nearly half of the model size and higher inference speed. When we reduce the network dimension of POMO$^\#$ to 64 (as our AMDKD), POMO$^\#$ (64) performs much inferior to our AMDKD-POMO (64).}
Pertaining to CVRP, similar patterns can be observed, where our AMDKD-AM (64) and AMDKD-POMO (64) consistently deliver favourable generalization performance against all the compared baselines.
{Besides, both our AMDKD-AM (64) and AMDKD-POMO (64) exhibit significantly faster inference than improvement methods LCP and DACT while achieving similar or even lower gaps.}

Furthermore, we show that by coupling our AMDKD-POMO (64) model with the recent \emph{efficient active search} (EAS)~\cite{eas}, the resulting AMDKD+EAS (\textit{T}=100) attains considerably superior performance that even surpasses LKH3 on CVRP-100 with much shorter time. Given the advances of active search for a particular instance to be inferred, AMDKD+EAS allows the model learned by our AMDKD with strong generalization to continuously improve its performance during inference (longer \textit{T} will yield even better performance), which leads to a new state-of-the-art hybrid solver.

We now evaluate the generalization of our AMDKD on the benchmark, i.e., TSPLIB and CVRPLIB,  which contain various instances of unknown distributions and larger size.
As shown in Table \ref{tab:libsummary}, our AMDKD-AM (64) significantly boosts the generalization of AM (128) and also outperforms the baseline AM$^\#$ (128) as well as other generalization methods that take AM (128) as the backbone model (GANCO and HAC). Similarly, our AMDKD-POMO (64) also significantly beats the other baseline methods, {except for POMO$^\#$ (128) on TSPLIB. Nevertheless, regarding the harder CVRP instances from CVRPLIB, our AMDKD-POMO (64) could yield significantly better performance.}  
Finally, we note that the AMDKD+EAS achieves the smallest gaps for both TSP and CVRP among all baselines. We refer to Appendix~\ref{app:lib} for full results
and discussions.

\begin{table}[t]
\centering
\caption{Generalization performance on selected instances (100 $\!\leq\!n\!\leq\!$ 200) from benchmark datasets.}\vspace{6pt}
\label{tab:libsummary}
  \resizebox{0.99\textwidth}{!}{
\begin{tabular}{@{}r|c|ccccc|ccccc@{}}
\toprule
& PSRO & AM (128)  & GANCO     & HAC   & AM$^\# (128) $& \textbf{AMDKD-AM} (64) & POMO (128)   & DROP   & POMO$^\# (128) $& \textbf{AMDKD-POMO} (64) & \textbf{AMDKD+EAS} \\ \midrule
TSPLIB  & 4.47\% & 42.63\% & 4.87\%    & 6.06\% & 17.60\% & \textbf{3.53\%}
& 29.73\% & 10.79\% & {\textbf{0.87\%}} & {1.08\%}   & \textbf{{0.74\%}}  \\
CVRPLIB & - & 29.36\% &  -    & - & 13.88\% & \textbf{7.43\%} & 14.19\% & 8.67\% & 6.80\% & \textbf{4.38\%} & \textbf{{1.26\%}}     \\ \bottomrule
\end{tabular}
}
\end{table}
\subsection{Further analysis of AMDKD}
\label{sec:ablation_md}

\textbf{Effects of different components.}  
We conduct ablation studies on CVRP-50 and TSP-50 to verify the designs of our AMDKD, where we consider removing the proposed adaptive strategy, the KD loss $\mathcal{L}_\text{KD}$, and the task loss $\mathcal{L}_\text{Task}$, respectively. 
As displayed in Table \ref{tab:ablation}, the use of adaptive strategy further enhances the learning effectiveness, and the two losses play a prominent role in the phase of gradient update. We also verify that when compared with our on-policy scheme, the off-policy variant (as mentioned in Section \ref{sec:adaptive}) saliently impairs the distillation performance.
\begin{table}
	\begin{minipage}{0.55\linewidth}
	\vspace{-5pt}
		\caption{{Ablation studies of AMDKD designs.}}
		\vspace{6pt}
		\label{tab:ablation}
		\centering
  \resizebox{0.9\textwidth}{!}{
    \begin{tabular}{cccccrr}
    \toprule
    \multirow{2}[4]{*}{Model} & \multicolumn{4}{c}{Components} &\multicolumn{1}{c}{CVRP-50}   & \multicolumn{1}{c}{TSP-50} \\
 &  $p^\text{adaptive}$ &  $\mathcal{L}_\text{KD}$ & \multicolumn{1}{c}{$\mathcal{L}_\text{Task}$}& $\tau_{\theta^\text{S}}$  & \multicolumn{1}{c}{Avg. Gap} & \multicolumn{1}{c}{Avg. Gap} \\
    \midrule
     \textit{w/o} adaptive &  & \checkmark    & \checkmark    & \checkmark  &  0.94\% & 0.69\textperthousand \\
    \textit{w/o} $\mathcal{L}_\text{KD}$ & \checkmark     &  
    & \checkmark   & \checkmark   & 1.06\% & 0.66\textperthousand \\
    \textit{w/o} $\mathcal{L}_\text{Task}$  & \checkmark     & \checkmark    &     &\checkmark  & 1.00\% & 0.69\textperthousand \\
    off-policy & \checkmark     & \checkmark    & \checkmark    &   & 10.42\% &  0.80\textperthousand \\
    \midrule
    AMDKD & \checkmark     & \checkmark    & \checkmark    & \checkmark & \textbf{{0.90\%}} &  \textbf{0.63\textperthousand}\\
    \bottomrule
    \end{tabular}%
    }\vspace{-10pt}
	\end{minipage}
	\hfill
	\begin{minipage}{0.45\linewidth}
		\centering
		\subfloat[]{\includegraphics[width = 0.43\textwidth]{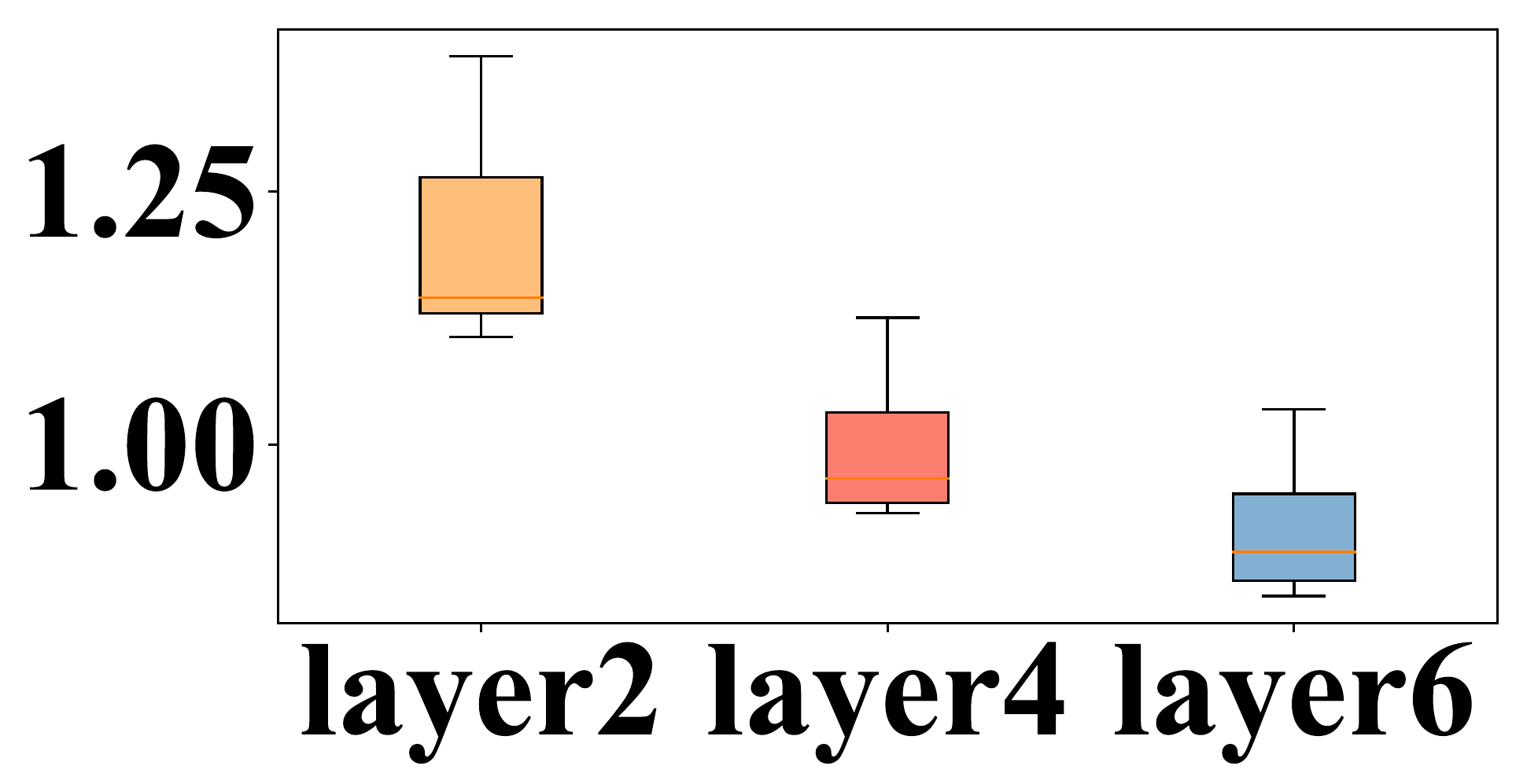}}
     \hspace{1pt}
         \subfloat[]{\includegraphics[width = 0.43\textwidth]{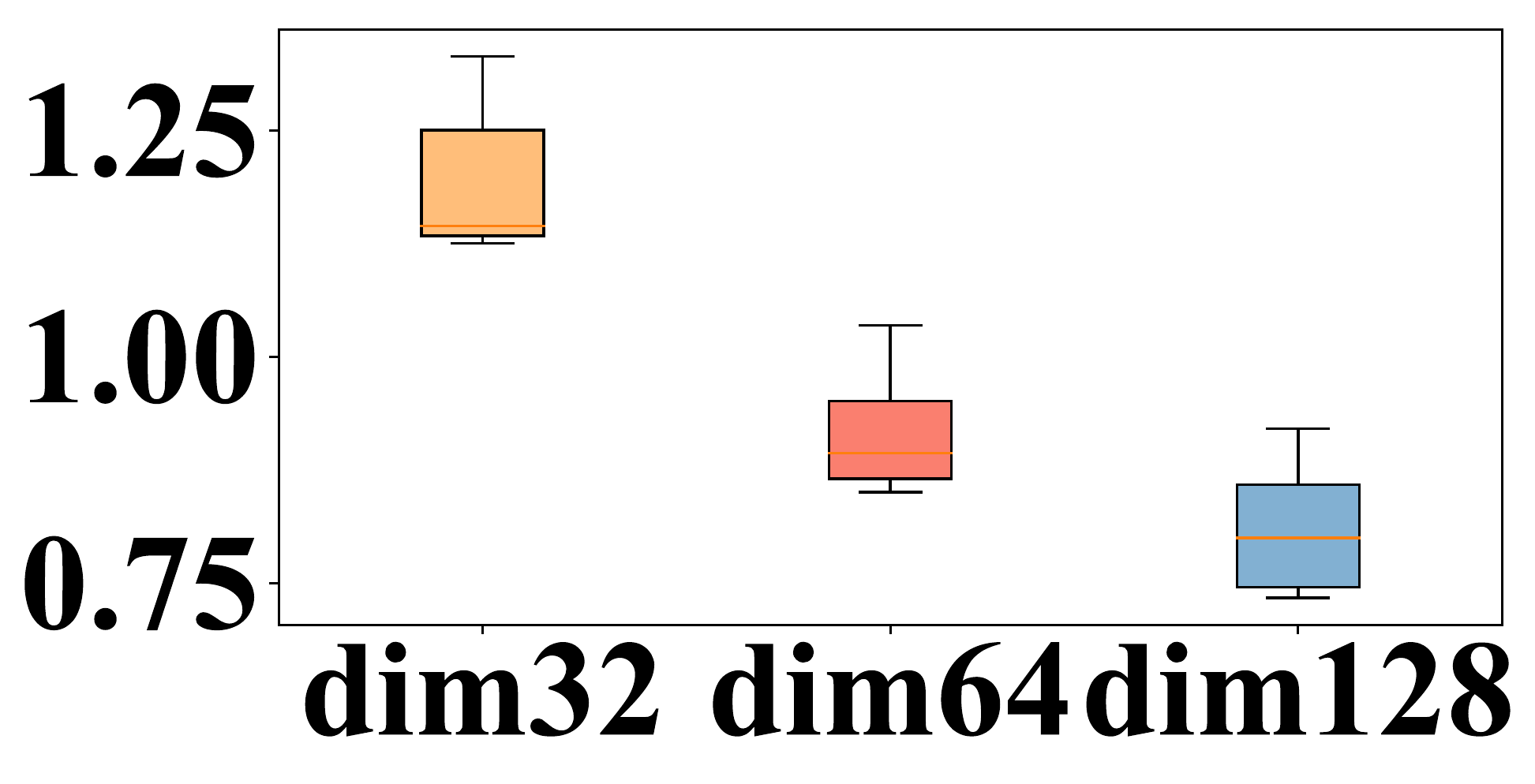}}
     \captionof{figure}{Effects of student network architectures. (a) different numbers of encoder layers, (b) different node embedding dimensions.}
     \label{figu:re}
     \vspace{-10pt}
	\end{minipage}
\end{table}

\textbf{Effects of student network architectures.}  
In Figure \ref{figu:re}, we exhibit the boxplots of the averaged gaps when using different student network architectures (different dimensions of the entire network and different number of layers in its encoder) of AMDKD-POMO for solving CVRP-50. As revealed, the larger the model, the better its performance. Hence, we acknowledge that our AMDKD can be further boosted when the student (64) shares the same architecture with its teachers (128). {Specifically, for CVRP-50, our AMDKD-POMO (128) exhibits an average optimality gap of 0.81\%, which is lower than the 0.90\% of AMDKD-POMO (64). However, this improvement comes at the expense of increased computational costs in three aspects:  1) more training time - AMDKD-POMO (128) needs almost 1.7 times (12 days vs 7 days) more training time; 2) more parameters - AMDKD-POMO (128) has about 2.4 times (1.20 M vs 0.49 M) more parameters; 3) slower inference time - AMDKD-POMO (128) infers about 1.2 times (1.3 min vs 1.1 min) slower than AMDKD-POMO (64). Hence, we note that there is a trade-off between solution quality and computational cost.}

{
\textbf{Effects of different exemplar distributions.} 
We now vary the distributions which were used as the exemplar ones to validate that AMDKD could still be effective. We take the POMO and CVRP-50 as an example.
To distinguish from the original version (AMDKD-POMO), we term the model taking Expansion, Implosion, Explosion, and Grid as the exemplar distributions as AMDKD-POMO*, and gathered the results in Table \ref{tab:exemplar}.
It shows that the overall performance of AMDKD-POMO* is still better than POMO  trained on these four distributions respectively. Hence, our AMDKD is able to consistently improve the cross-distribution generalization with different exemplar distributions.}

\begin{table}[tb]
  \centering
  \caption{Performance of AMDKD on CVRP-50 trained with different exemplar distributions.}
  \label{tab:exemplar}
  \vspace{5pt}
    \resizebox{0.99\textwidth}{!}{
    \begin{tabular}{l|c|cc|cc|cc|cc|cc|cc|cc|c}
\toprule
\multirow{2}[2]{*}{Model} & \multicolumn{1}{c|}{Size} & \multicolumn{2}{c|}{Expansion} & \multicolumn{2}{c|}{Implosion} & \multicolumn{2}{c|}{Explosion} & \multicolumn{2}{c|}{Grid} & \multicolumn{2}{c|}{Uniform} & \multicolumn{2}{c|}{Cluster} & \multicolumn{2}{c|}{Mixed} & \multicolumn{1}{c}{\multirow{2}[2]{*}{Avg. Gap}} \\
    \multicolumn{1}{c|}{} & \multicolumn{1}{c|}{(M)} & \multicolumn{1}{c}{Obj.} & Gap   & \multicolumn{1}{c}{Obj.} & Gap   & \multicolumn{1}{c}{Obj.} & Gap   & \multicolumn{1}{c}{Obj.} & Gap   & \multicolumn{1}{c}{Obj.} & Gap   & \multicolumn{1}{c}{Obj.} & Gap   & \multicolumn{1}{c}{Obj.} & Gap   &  \\
    \midrule
    LKH3  & \multicolumn{1}{c|}{-} & 8.15  & - & 10.26  & -& 8.74  & - & 10.40  & - & 10.38  & - & 5.13  &-  & 9.42  & - & - \\
    POMO(Expansion) & 1.20  & 8.22  & 0.90\% & 10.36  & 0.96\% & 8.82  & 0.94\% & 10.50  & 0.95\% & 10.47  & 0.94\% & 5.19  & 1.14\% & 9.51  & 0.96\% & 0.97\% \\
    POMO(Implosion) & 1.20  & 8.23  & 0.96\% & 10.34  & 0.80\% & 8.81  & 0.82\% & 10.48  & 0.77\% & 10.46  & 0.78\% & 5.21  & 1.50\% & 9.52  & 1.03\% & 0.95\% \\
    POMO(Explosion) & 1.20  & 8.23  & 0.95\% & 10.35  & 0.83\% & 8.81  & 0.80\% & 10.49  & 0.82\% & 10.47  & 0.87\% & 5.20  & 1.27\% & 9.51  & 1.02\% & 0.94\% \\
    POMO(Grid) & 1.20  & 8.23  & 0.97\% & 10.34  & 0.79\% & 8.81  & 0.82\% & 10.48  & 0.76\% & 10.46  & 0.77\% & 5.21  & 1.57\% & 9.52  & 1.07\% & 0.97\% \\
    \midrule
    AMDKD-POMO* & 0.49  & 8.23  & 0.96\% & 10.35  & 0.89\% & 8.82  & 0.89\% & 10.49  & 0.88\% & 10.47  & 0.88\% & 5.18  & 1.06\% & 9.50  & 0.92\% & \textbf{0.92\%} \\
    \bottomrule
    \end{tabular}%
    }
\end{table}%

\section{Conclusions and future work}
\label{sec:conclusion}
In this paper, we propose an Adaptive Multi-Distribution Knowledge Distillation (AMDKD) scheme to alleviate the cross-distribution generalization issue of deep models for VRPs.
Different from the existing generalization methods, we aim to enhance the distribution generalization by transferring various policies learned from exemplar distributions into one via an efficient knowledge distillation scheme. To facilitate effective learning, we design an adaptive strategy to train a single yet generalist student network by leveraging multiple teachers in turns. 
The experiment results exhibit competitive performance of our AMDKD in generalizing to other unseen out-of-distribution instances (randomly generated or from benchmarks), which also consumes less computational resources. While our AMDKD is generic, a potential limitation is that its boost is not guaranteed to be always significant across all unseen distributions for all backbone models. 
For future work, we will investigate, 1) {generalizing AMDKD for different/larger problem sizes}; 2) considering the \emph{improvement} models like DACT~\cite{ma2021learning} as the backbone; 3) performing online distillation to jointly and efficiently train the teachers and the student models~\cite{guo2020online}; {
4) assessing the impact of the quality of the validation dataset on the distillation;
and 5) enhancing the interpretability of AMDKD \cite{zhang2022quantifying}.}

\begin{ack}
{This work is supported in part by the National Key R\&D Program of China (2018AAA0101203), the National Natural Science Foundation of China (62072483), and the Guangdong Basic and Applied Basic Research Foundation (2022A1515011690, 2021A1515012298); in part by the Agency for Science Technology and Research Career Development Fund (C222812027), and the IEO Decentralised GAP project of Continuous Last-Mile Logistics (CLML) at SIMTech (I22D1AG003).}
\end{ack}

{
\small
\bibliographystyle{unsrtnat}
\bibliography{main}
}
\clearpage

\appendix
\setcounter{page}{1} 
\appendix

\vbox{
\hrule height 4pt
\vskip 0.25in
\vskip -\parskip%
\centering
{\LARGE\bf Learning Generalizable Models for Vehicle Routing
Problems via Knowledge Distillation (Appendix)}
\vskip 0.29in
\vskip -\parskip
\hrule height 1pt
}

\section{Details of the considered distributions}
\label{app:dist}
In this paper, we consider various distributions for the node coordinates in VRPs, followed which we randomly generate instances for both training and testing. Below we present details on how to generate those instances. Specifically, we follow the recent work~\cite{kool2018attention} and benchmark dataset TSPLIB~\cite{reinelt1991tsplib} to generate instances of Uniform, Cluster and Mixed distributions, and follow the settings in \cite{Jiang2022LearningTS,Bossek2019EvolvingDT} for the Expansion, Implosion, Explosion, and Grid ones.

\textbf{Uniform distribution.}
It considers uniformly distributed nodes. Following~\cite{kool2018attention}, we generate the two-dimensional coordinates $(x, y)$ of each node by sampling from a uniform space $\mathcal{U}$([0, 1]$^2$). An exemplary instance is displayed in Figure~\ref{fig:dist}(h).

\textbf{Cluster distribution.} 
It considers multiple ($n_c$) clusters, where we set {$n_c = 3$}. In specific, each cluster follows a normal distribution $\mathcal{N}$($\mu$, $\sigma^2$)$^2$ with the mean sampled uniformly, i.e., $\mu \sim \mathcal{U}$([0.2, 0.8]$^2$) and the standard deviation $\sigma$ = 0.07.
According to the 3$\sigma$ rule, each node has a 99.7\% probability of being generated in the [0, 1]$^2$ square region, and outliers will have their coordinates re-modified, where values less than 0 are changed to 0 and those greater than 1 are changed to 1, to ensure all coordinates are constrained to ([0, 1]$^2$). An exemplary instance is displayed in Figure~\ref{fig:dist}(i).

\textbf{Mixed distribution.} 
It considers a mixture of the two distributions above, each with half of the nodes. For the latter, we only consider $n_{c}\!=\!1$ cluster. An exemplary instance is displayed in Figure~\ref{fig:dist}(j).

\textbf{Expansion distribution.}
It considers a linear function to mutate the nodes in Uniform distribution. Gvien a randomly generated linear function $y = ax + b$, all nodes, orthogonal to the linear function within the distance $r$ ($r = 0.3$), are moved away from their original coordinates to a farther position, whose orthogonal distance is $r+\gamma$, where $\gamma$ obeys an exponential distribution with the rate parameter $\lambda = 10$, i.e. $\gamma \sim E (\lambda)$. Regarding the linear function, we first sample $b$ (intercept) from [0, 1], then $a$ (slope) is sampled uniformly from [0, 3] (if $b$ < 0.5) or [-3, 0] (if $b \geq$ 0.5). Finally, we normalize all node coordinates $X$ as follows to ensure that they are constrained to [0, 1]$^2$,
\begin{equation}
    X' = \frac{X-\text{min}(X)}{\text{max}(X)-\text{min}(X)},
    \label{eq:norm}
\end{equation}
where $X'$ denotes the normalized coordinates. An exemplary instance is displayed in Figure~\ref{fig:dist}(k).

\textbf{Implosion distribution.} 
It considers an implosion to mutate the nodes in Uniform distribution.
To simulate an implosion, it first samples a centroid $\epsilon_\text{i}$, then gathers all nodes within the circle of $\epsilon_\text{i}$ (with the radius $R_\text{ic} = 0.3$) together towards a new circle with the same centroid $\epsilon_\text{i}$ but a (randomly sampled) smaller radius ($R_\text{i} \leq 0.3$).  
An exemplary instance is displayed in Figure~\ref{fig:dist}(l).

\textbf{Explosion distribution.} 
It considers to mutate the nodes in Uniform distribution by imitating the particles affected by an explosion. Similar to Implosion, it first randomly samples a centroid. Then, instead of gathering all nodes towards the centroid in Implosion distribution, it moves away those nodes from the circle (radius $R_\text{ec}\!=\!0.3$) and explode them outside the circle, which follow the direction vector between the centroid $\epsilon_\text{e}$ and the corresponding nodes. The additive distance $\gamma$ is randomly sampled from an exponential distribution with a rate parameter, i.e. $\gamma \sim E (\lambda)$. All nodes are them normalized using Eq. (\ref{eq:norm}). An exemplary instance is displayed in Figure~\ref{fig:dist}(m).

\textbf{Grid distribution.}
It considers to mutate the nodes in Uniform distribution by imposing a grid permutation. We first generate the four vertex of a square with the width and height equal to $R_\text{g} (R_\text{g} = 0.3)$ and then place it within the region [0, 1]$^2$. All the pre-generated nodes inside the box are re-arranged as a quadratic grid instead. In this case, all nodes are constrained to [0, 1]$^2$. An exemplary instance is displayed in Figure~\ref{fig:dist}(n).

The above distributions are considered in both TSP and CVRP. For extra settings in CVRP, we follow the convention~\cite{kool2018attention,ma2021learning,kwon2020pomo}. In specific, the demand $\delta_i$ of each node is sampled uniformly from $\mathcal{U}$(1, 2, $\cdots$, 9) and the capacity $Q$ of the vehicle varies with the problem scale, where we set $Q^{20}$ = 30, $Q^{50}$  = 40 and $Q^{100}$  = 50. For instances from CVRPLIB, we exactly follow their settings.

\section{Details of compared baselines}
\label{app:baseline}

\textbf{Implementation Details.}
We compare our AMDKD with various types of baselines. Regarding the neural baselines, we re-train LCP~\cite{Kim2021LearningCP}, HAC~\cite{Zhang2022LearningTS} and DACT~\cite{ma2021learning} on our machine based on the code that are publicly available on Github. By default, we follow their original settings and the suggestion on the hyper-parameters. More details of the baselines are presented below.
\begin{itemize}
\setlength\itemsep{0em}
    \item \textbf{Gurobi} \cite{gurobi}: we use Gurobi to obtain the optimal solutions to TSP instances, which are implemented under the default settings.
    \item \textbf{LKH} \cite{lkh3}: For CVRP, it is usually hard to obtain the optimal solutions. Thus, we use the strong LKH solver to find near-optimal solutions. Note that the LKH solver is also a widely used baseline to evaluate and compare the recent learning based methods for VRPs, where we run it following the conventions in   \cite{kool2018attention,ma2021learning,kwon2020pomo,eas}.  
    \item \textbf{LCP} \cite{Kim2021LearningCP}: LCP is a two-stage method, where a \emph{seeder} generates diverse initial solutions and a \emph{reviser} rewrites the current solutions partially. We re-train the LCP on a mixed dataset containing instances from the three exemplar distributions. For inference of TSP-50 and TSP-100, we employ the LCP* (the best version reported in~\cite{Kim2021LearningCP}) with two revisers (the lengths of the tour for revision are set to $\ell_{r1}$=10 and $\ell_{r2}$=20) and a sampling strategy (1,280), and set the total number of revision iteration $T_r$ to 45 (i.e., $T_{r1}$=25, $T_{r2}$=20, respectively). For TSP-20, we exploit one reviser ($\ell_r$=10) since the revision length must be less than the problem size according to its design, and set the number of revision iteration $T_r$ to 10.
    \item \textbf{HAC} \cite{Zhang2022LearningTS}: HAC designs a hardness-adaptive Gaussian instances generator to produce instances to fine-tune the given pre-trained AM model. In its original design, the dataset used for fine-tuning contains half instances uniformly distributed and the other half produced by its own generator. In this paper, we substitute the instances of uniform distribution with a mixed dataset containing instances from the three exemplar distributions.
    \item \textbf{DACT} \cite{ma2021learning}: DACT learns to guide the pairwise operator to perform local search. We adopt the 2-opt version since it reports the best result for TSP and CVRP according to its original paper~\cite{ma2021learning}. We re-train DACT on a mixed dataset containing instances from the three exemplar distributions, and set the iteration number to 1,280 for inference.
\end{itemize}

\begin{table}[b]
  \centering
  \caption{Effects of distribution mixture during training.}
  \vspace{10pt}
    \resizebox{0.99\textwidth}{!}{
    \begin{tabular}{cc|cccccccc|cccccccc}
    \toprule
    \multicolumn{2}{c|}{\multirow{2}[2]{*}{Model}} & \multicolumn{8}{c|}{{$n$ = 50}}                                   & \multicolumn{8}{c}{{$n$ = 100}} \\
    \multicolumn{2}{c|}{} & Uniform & Cluster & Mixed & Grid  & Implosion & Expansion & Explosion & Avg.  & Uniform & Cluster & Mixed & Grid  & Implosion & Expansion & Explosion & Avg. \\
    \midrule
    \multirow{11}[11]{*}{\rotatebox{90}{{TSP}}} & Gurobi & 5.70   & 2.65  & 4.92  & 5.69  & 5.60   & 4.38  & 4.62  & - & 7.76  & 3.66  & 6.73  & 7.79  & 7.61  & 5.39  & 5.83  & - \\
    \cmidrule{2-18}
          & POMO (U) & 5.70   & 2.66  & 4.93  & 5.69  & 5.60   & 4.38  & 4.62  & 0.12\% & 7.78  & 3.73  & 6.79  & 7.8   & 7.62  & 5.43  & 5.85  & 0.61\% \\
          & POMO$^\#$ & 5.70   & 2.66  & 4.93  & 5.69  & 5.60   & 4.38  & 4.62  & \textbf{0.06\%} & 7.78  & 3.67  & 6.75  & 7.81  & 7.63  & 5.42  & 5.84  & \textbf{0.29\% }\\
          \cmidrule{2-18}
          & AM (U) & 5.73  & 2.71  & 4.99  & 5.73  & 5.63  & 4.43  & 4.65  & \textbf{1.02\%} & 7.93  & 3.93  & 7.00     & 7.95  & 7.78  & 5.64  & 5.98  & 3.58\% \\
          & AM$^\#$  & 5.73  & 2.72  & 4.99  & 5.73  & 5.64  & 4.43  & 4.65  & 1.06\% & 7.92  & 3.90   & 7.00     & 7.95  & 7.77  & 5.66  & 5.98  & \textbf{3.40\%} \\
          \cmidrule{2-18}
          & LCP (U) &   5.72    &  3.58  &  5.01 &   5.71    &  5.62     &    4.44   &  4.65 &   5.70\%    &  7.95     &    6.91   &  7.34   &  7.98   &   7.81  &   6.23    & 6.26 & \textbf{18.31\%} \\
          & LCP $^\#$ & 5.73  & 3.11  & 5.04  & 5.72  & 5.63  & 4.47  & 4.65  & \textbf{3.44\%} & 7.99  & 6.81  & 7.34  & 8.02  & 7.85  & 6.41  & 6.30   & 18.72\% \\
        \cmidrule{2-18}
          & {HAC (U)} &   5.81 &	2.87 &	5.07  & 5.80 &	5.71 &	4.48 &	4.70  &  3.00\%  &   8.14 &	4.24 &	7.26  &  8.17 &	7.99 &	5.96 &	6.16   & 7.78\%  \\
          & {HAC$^\#$} & 5.72 &	2.70 &	4.97   & 5.72 &	5.63 &	4.41 &	4.64 &  \textbf{0.78\%} & 7.96 &	3.94 &	6.98 & 7.98 &	7.81 &	5.60 &	6.01 & \textbf{3.70\%}  \\
          \cmidrule{2-18}
          & DACT (U) &   5.70    &  2.69     & 4.98      &   5.70  & 5.61    &  4.43     &  4.63    & 0.61\%  &   7.90    &  3.90  &  6.98 &   7.92   &  7.75     & 5.80    &      5.99 & 3.67\%  \\
          & DACT$^\#$ & 5.71  & 2.66  & 4.94  & 5.71  & 5.62  & 4.40   & 4.63  & \textbf{0.33\%} & 7.96  & 3.74  & 6.88  & 7.98  & 7.80   & 5.67  & 5.98  & \textbf{2.80\%} \\
    \midrule
    \multirow{7}[7]{*}{\rotatebox{90}{{CVRP}}} & LKH3  & 10.38 & 5.13  & 9.42  & 10.40  & 10.26 & 8.15  & 8.74  & - & 15.65 & 7.81  & 14.19 & 15.64 & 15.44 & 11.39 & 12.32 & - \\
    \cmidrule{2-18}
          & POMO (U) & 10.46 & 5.21  & 9.52  & 10.49 & 10.35 & 8.23  & 8.81  & 0.98\% & 15.80  & 7.99  & 14.38 & 15.87 & 15.59 & 11.55 & 12.45 & 1.36\% \\
          & POMO $^\#$ & 10.47 & 5.18  & 9.50   & 10.50  & 10.36 & 8.23  & 8.82  & \textbf{0.93\%} & 15.83 & 7.91  & 14.32 & 15.82 & 15.62 & 11.55 & 12.46 & \textbf{1.18\% }\\
          \cmidrule{2-18}
          & AM (U) & 10.64 & 5.35  & 9.70   & 10.66 & 10.52 & 8.39  & 8.96  & \textbf{2.92\%} & 16.13 & 8.58  & 14.84 & 16.13 & 15.93 & 11.93 & 12.77 & 4.59\% \\
          & AM$^\#$  & 10.63 & 5.37  & 9.71  & 10.66 & 10.52 & 8.40   & 8.97  & 2.97\% & 16.16 & 8.46  & 14.85 & 16.15 & 15.95 & 11.93 & 12.78 & \textbf{4.47\%} \\
          \cmidrule{2-18}
        & DACT (U) & {10.54} & {5.24}  & {9.57}  & {10.56} & {10.42} & {8.26}   & {8.88}  & {1.62\%} & {16.15} & {8.17}  & {14.74}  & {16.14} & {15.94} & {12.33} & {12.74} & {\textbf{4.22\%} }\\
        & DACT$^\#$ & {10.54} & {5.21}  & {9.57}  & {10.57} & {10.43} & {8.27}   & {8.89}  & {\textbf{1.59\%}} & {16.52} & {8.25}  & {14.97}  & {16.51} & {16.30} & {12.02} & {13.01} & {{5.56\%} }\\
    \bottomrule
    \end{tabular}%
    }
  \label{tab:mixdata}%
\end{table}%

\textbf{Effects of distribution mixture augmentation.}
To ensure fair comparisons, we re-train the baselines on a mixed dataset containing instances from the three exemplar distributions (with $^\#$). However, we notice that this simple distribution augmentation does not always lead to a better generalization, espicially for the methods that have an \emph{improvement} component. {For example, regarding the \emph{improvement} method DACT, we find that DACT$^\#$ performs even worse than DACT trained on Uniform distribution for CVRP-100; and regarding the hybrid method, LCP (U) performs slightly better than LCP$^\#$ on TSP-100.} Nevertheless, this distribution augmentation improves the generalization of construction methods POMO \cite{kwon2020pomo} and AM \cite{kool2018attention} on larger instances, i.e., TSP-100 and CVRP-100.

{
\textbf{Why AMDKD could be better than distribution mixture augmentation?} Recall that we conclude from Table~\ref{tab1} that our AMDKD usually achieves better cross-distribution generalization performance when compared to the above distribution mixture method (\# models). Though this still remains a open question, we list some possible intuitions on why AMDKD works better as follow:
\begin{itemize}
\item Different exemplar distributions may have different levels of difficulty for solving. Without additional intervention, reinforcement learning tends to learn those easier things to get a higher reward. This means that training on mixed data may possibly bias the learning towards distributions that are easier to solve (a "winner-take-all" issue). On the contrary, our AMDKD selects a specific teacher model based on the weakness of the current student model (by our adaptive strategy), which encourages it to learn those hard-to-solve distributions. And this adaptive strategy echoes how humans learn knowledge, where more time is always required for harder subjects.
\item Directly training on a mixed dataset may not be efficient or stable. On one hand, it would be hard for deep reinforcement learning to directly learn good patterns from mixed data that follow different distributions, due to the possibly limited representation capability of the neural networks for handling such hard optimization problems, even without the diversity in distributions. On the other hand, the tasks for different distributions may have different reward ranges (such as the route length), which may cause instability in the RL training. For example, the average total rewards for solving CVRP-100 instances following Uniform and Cluster distributions are around 16 and 7, respectively. Our AMDKD tackles this issue in a way that only one distribution is leveraged in a training epoch. Further empowered by the knowledge distillation, our AMDKD framework efficiently and effectively transfers useful knowledge (patterns) from various teacher models to a unified and light student model.
\end{itemize}
}

\section{Additional analysis of AMDKD}
\label{app:ablation}

\subsection{Effects of multiple teacher co-training.}
\begin{figure}[H]
    \centering
    \includegraphics[width=0.13\textwidth]{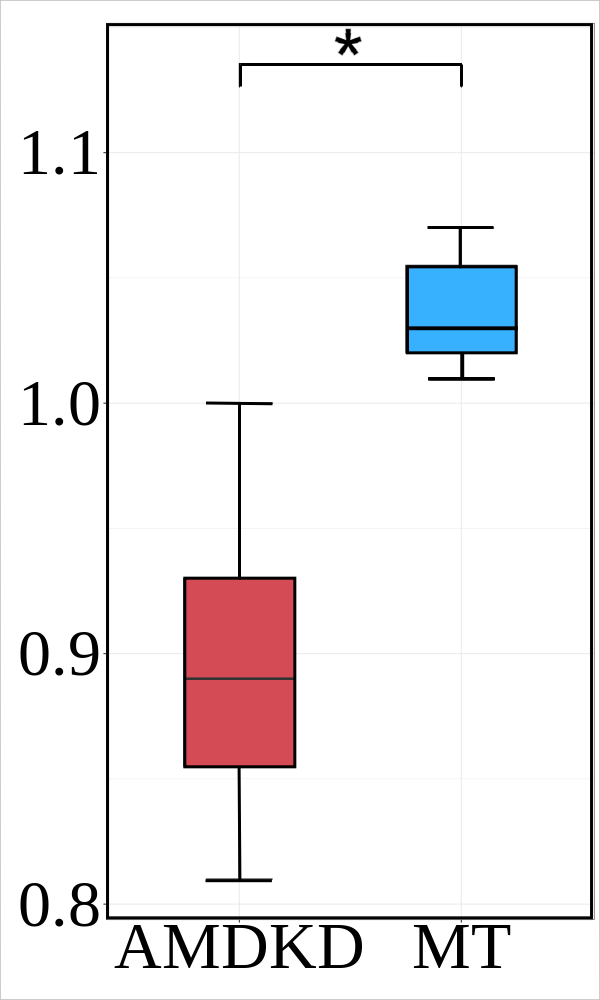}
    \caption{Boxplot of the overall gaps of AMDKD and its MT version. Here, $\star$ in the plot means that the two models are much different with statistical significance $p$-value = 0.02 < 0.05 (Wilcoxon test).}
    \label{fig:multi}
\end{figure}

Recall that AMDKD selects only one distribution and its corresponding teacher in each epoch. Different from ours, some existing multi-teacher knowledge distillation approaches exploit multiple teachers simultaneously in each epoch. We term such strategy as MT, whose loss function follows Eq. (\ref{eq:multi}). In Figure \ref{fig:multi}, we draw the boxplot of the overall gaps (on all seven test distributions for CVRP-50) of AMDKD-POMO and its MT version, and compare the results using the Wilcoxon test. As clearly demonstrated, allowing unprofessional teachers to advise for distributions in which they are not specialized will interfere with the process of knowledge distillation, inducing significantly inferior performance compared to ours.

\begin{figure}[H]
    \centering
    \setlength{\abovecaptionskip}{0.3cm} 
   \setlength{\belowcaptionskip}{-0.3cm}
    \includegraphics[width=0.13\textwidth]{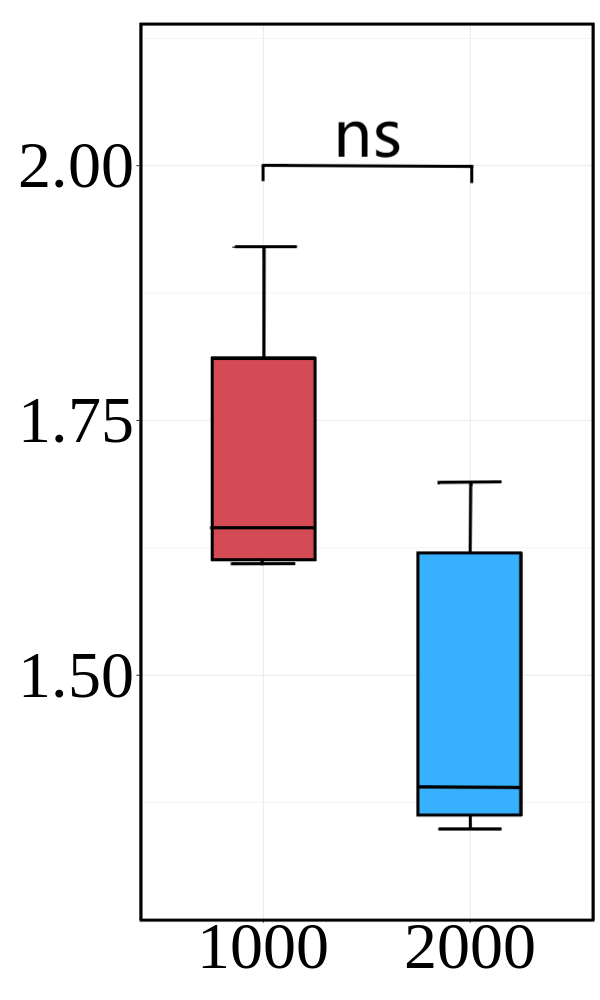}
    \caption{{Boxplot of the overall gaps of AMDKD-AM (with size of $\mathcal{V}$=1,000, red) and AMDKD-AM (with size of $\mathcal{V}$=2,000, blue) on CVRP-50. The "ns" in the plot means that the two models are not sigificantly different with statistical significance $p$-value = 0.25 $>$ 0.1 (Wilcoxon test).}}
    \label{fig:valid}
\end{figure}

\begin{figure}[H]
    \centering
    \setlength{\abovecaptionskip}{0.3cm} 
   \setlength{\belowcaptionskip}{-0.3cm}
    \includegraphics[width=0.13\textwidth]{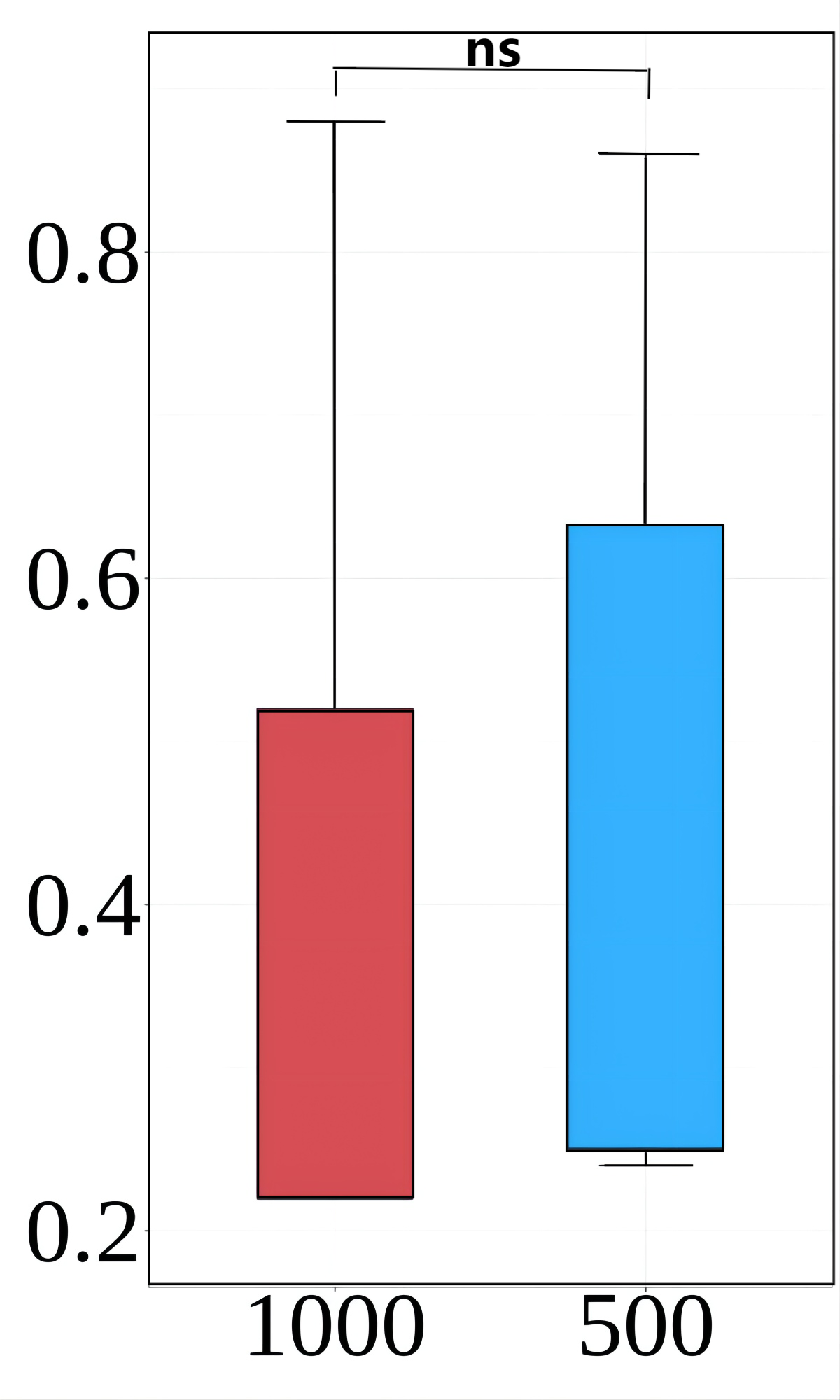}
    \caption{{Boxplot of the overall gaps of AMDKD-AM (with size of $\mathcal{V}$=1,000, red) and AMDKD-AM (with size of $\mathcal{V}$=500, blue) on TSP-50. The "ns" in the plot means that the two models are not sigificantly different with statistical significance $p$-value = 0.36 $>$ 0.1 (Wilcoxon test).}}
    \label{fig:valid500}
\end{figure}

\begin{figure}[H]
    \centering
    \includegraphics[width=0.5\textwidth]{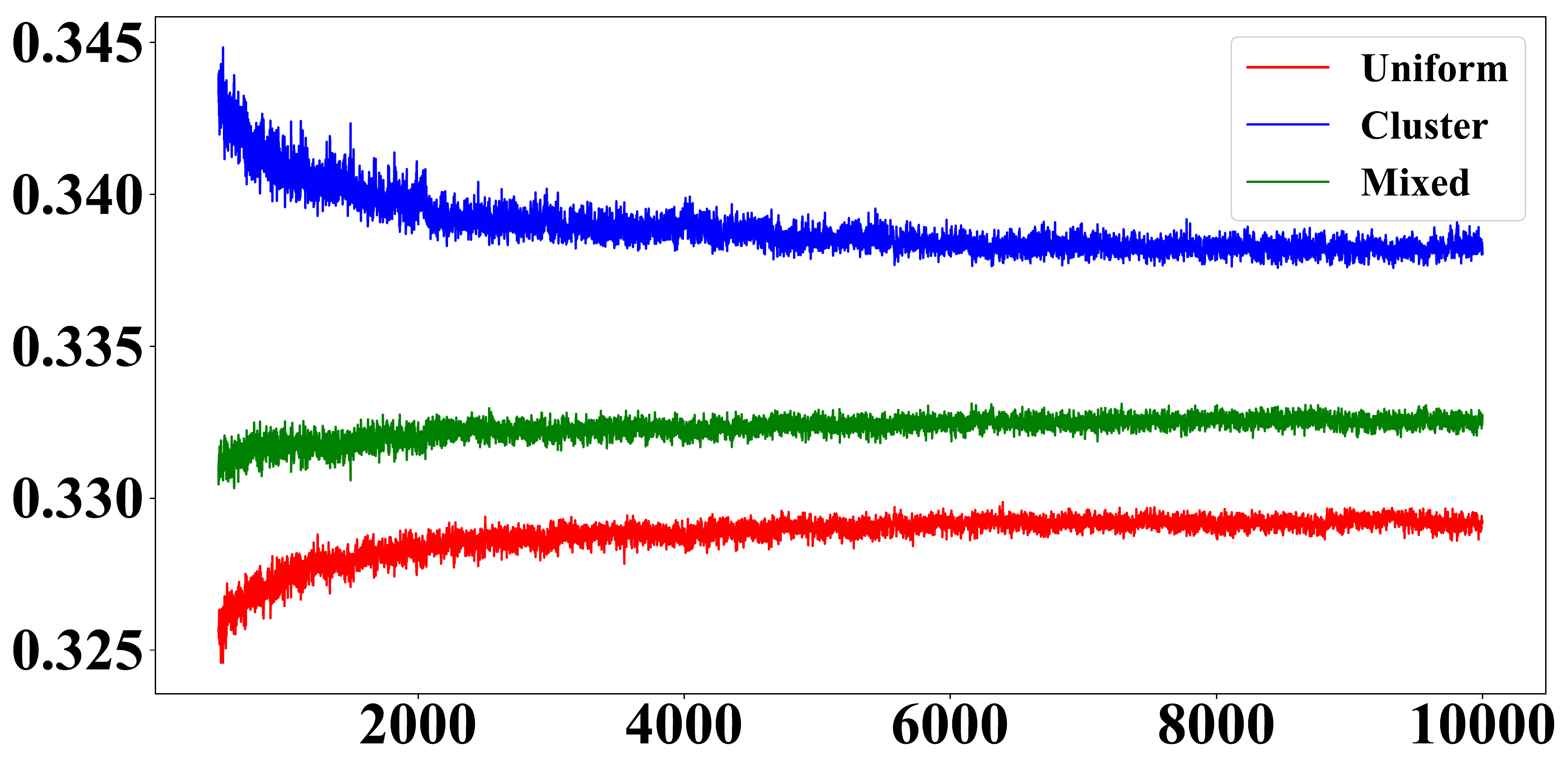}
    \vspace{-6pt}
    \caption{Likelihood of teacher selection along the epochs (AMDKD-AM on CVRP-100).}
    \label{fig:likelihood}
\end{figure}

{\subsection{Effects of validation datasets.} Recall that the adaptive probability of teacher selection in Eq. (\ref{eq:softmax}) is calculated by the real-time performance of student model on the validation datasets, we further investigate whether the size of the validation datasets $\mathcal{V}$ will largely influence AMDKD. As displayed in Figure \ref{fig:valid} and Figure \ref{fig:valid500}, increasing the size of $\mathcal{V}$ (from 1,000 to 2,000) slightly improves the performance of AMDKD (but with no statistical significance), whereas decreasing the size of $\mathcal{V}$ (from 1,000 to 500) slightly impairs the performance of AMDKD (also with no statistical significance), which indicates that the good performance of AMDKD may not rely on the size of the validation datasets. Meanwhile, performing student model evaluation in each epoch inevitably introduces additional computation cost, where larger size of the validation datasets will cause longer training time. However, we note that the increment of validation in training time is acceptable when we use $\mathcal{V}$ = 1,000. Taking AMDKD-AM training on CVRP-100 as an example, the total evaluation time is 2.7s (0.9s per exemplar distribution on average) for each epoch, which is approximately 1\% of the total training time (i.e., 4 min). What's more, we note that the likelihood of selecting different teacher models would eventually converge to a stable one, which means that we may stop such evaluation early to speed up the training further if needed. For example, for training AMDKD-AM on CVRP-100 (see Figure \ref{fig:likelihood}), we may stop the student evaluation early at around 3,000 epochs.}

\begin{figure}[H]
    \centering
    \setlength{\abovecaptionskip}{0.3cm} 
  \setlength{\belowcaptionskip}{-0.3cm}
    \includegraphics[width=0.13\textwidth]{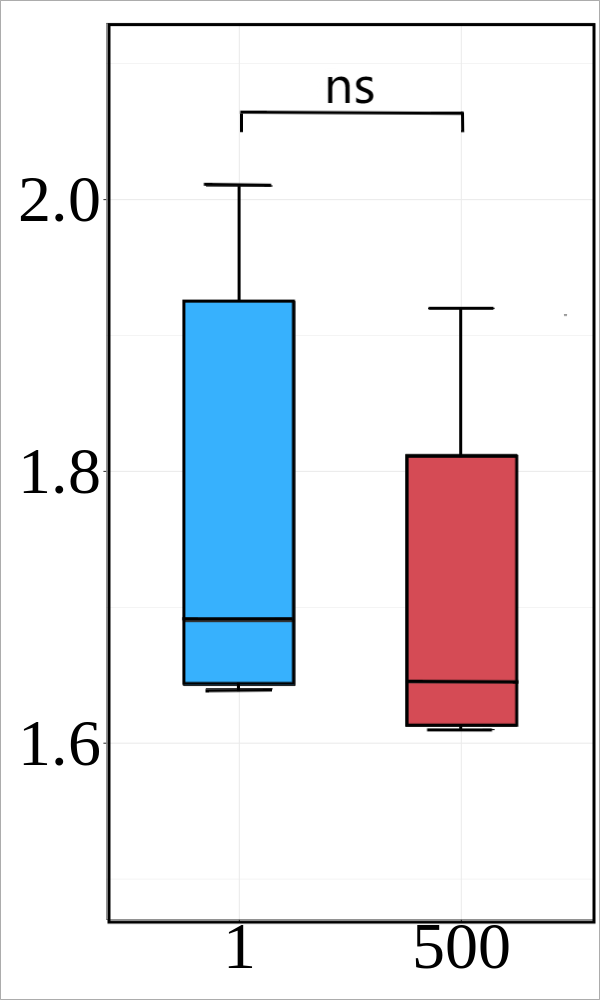}
    \caption{{Boxplot of the overall gaps of AMDKD-AM ($E'$=500, red) and AMDKD-AM ($E'$=1, blue) on CVRP-50. The "ns" in the plot means that the two models are not significantly different  with statistical significance $p$-value = 0.37 $>$ 0.1 (Wilcoxon test).}}
    \label{fig:e}
\end{figure}

{\subsection{Effects of different hyper-parameters.} We further discuss the influence of the hyper-parameters on the performance of AMDKD.
\begin{itemize}
    \item The starting epoch of the adaptive teacher selection strategy ($E'$): it indicates the epoch to start our adaptive strategy for teacher selection. We include this hyper-parameter because preliminary experiments revealed that there could be a point in the learning curve (i.e., the reward convergence curve) where the training curves without and with (starting at the first epoch $E'$=1) the adaptive strategy may meet. This suggests that there might be a sweet spot to implement the proposed adaptive strategy. For AMDKD-AM, the sweet spot is around $E'$=500. And for AMDKD-POMO, we do not observe such a pattern and thus we use $E'$)=1. In figure \ref{fig:e}, we provide an example of how $E'$ will affect the performance of AMDKD-AM.
    \item Number of steps per epoch ($T$): it indicates how long will the student model learn from the selected teacher before it possibly switches to a new one, which should not be too small or too large. As for POMO, the original $T$ is about 20, and we did not change it. As for AM, the original $T$ is about 2,500, and we empirically reduce it to 250 for a better trade-off.
    \item The total number of training epochs ($E$): it mainly follows the settings of the original backbone. In this paper, we employ different training epochs for different sizes and tasks since the hardness of the task itself may grow with size, which may require more steps to converge. {Training curves of AMDKD-POMO (for CVRP) are depicted in Figure \ref{fig:convergence}.
    }
\end{itemize}
}

\begin{table}[H]
{
	\begin{minipage}{\linewidth}
		\centering
		\subfloat[CVRP-20]{\includegraphics[width = 0.32\textwidth]{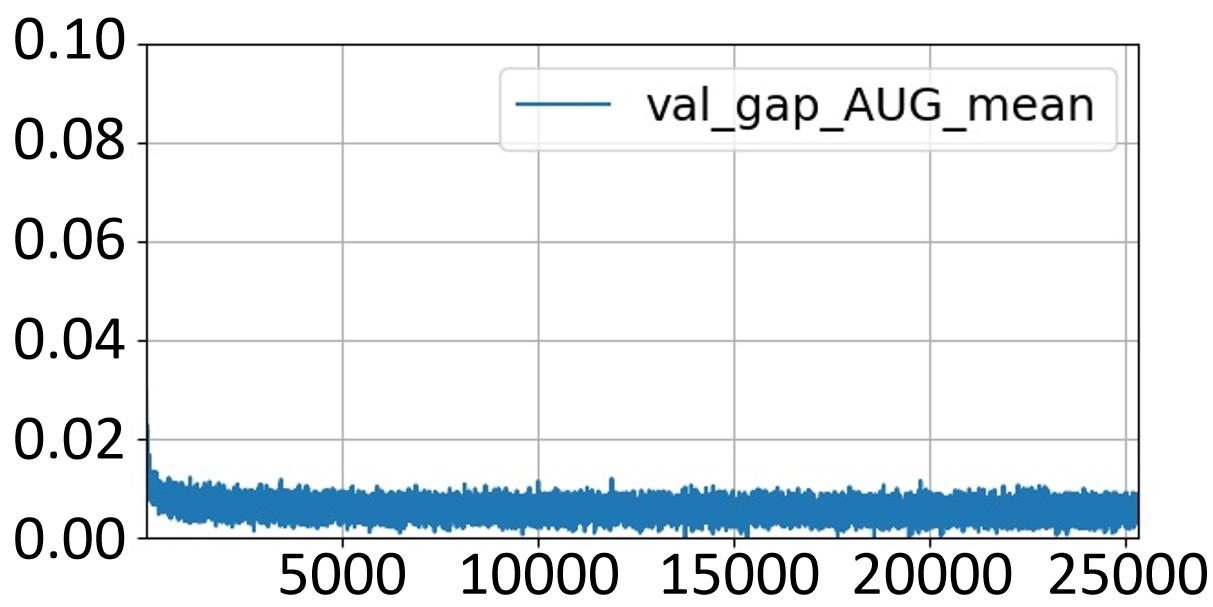}}
         \subfloat[CVRP-50]{\includegraphics[width = 0.32\textwidth]{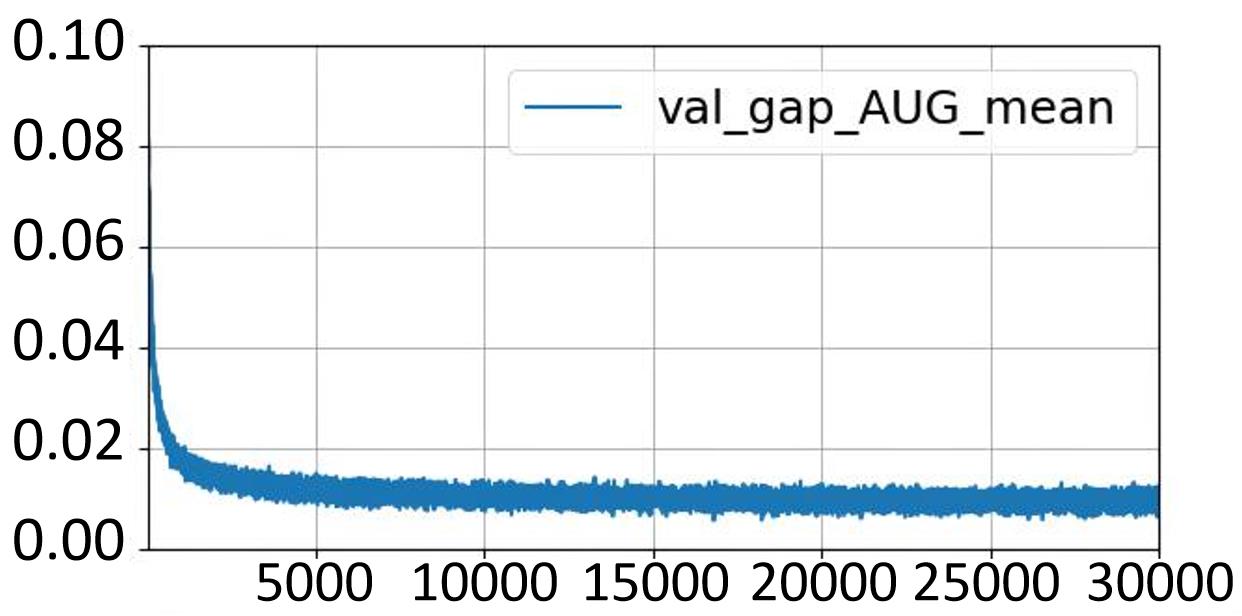}}
         \subfloat[CVRP-100]{\includegraphics[width = 0.32\textwidth]{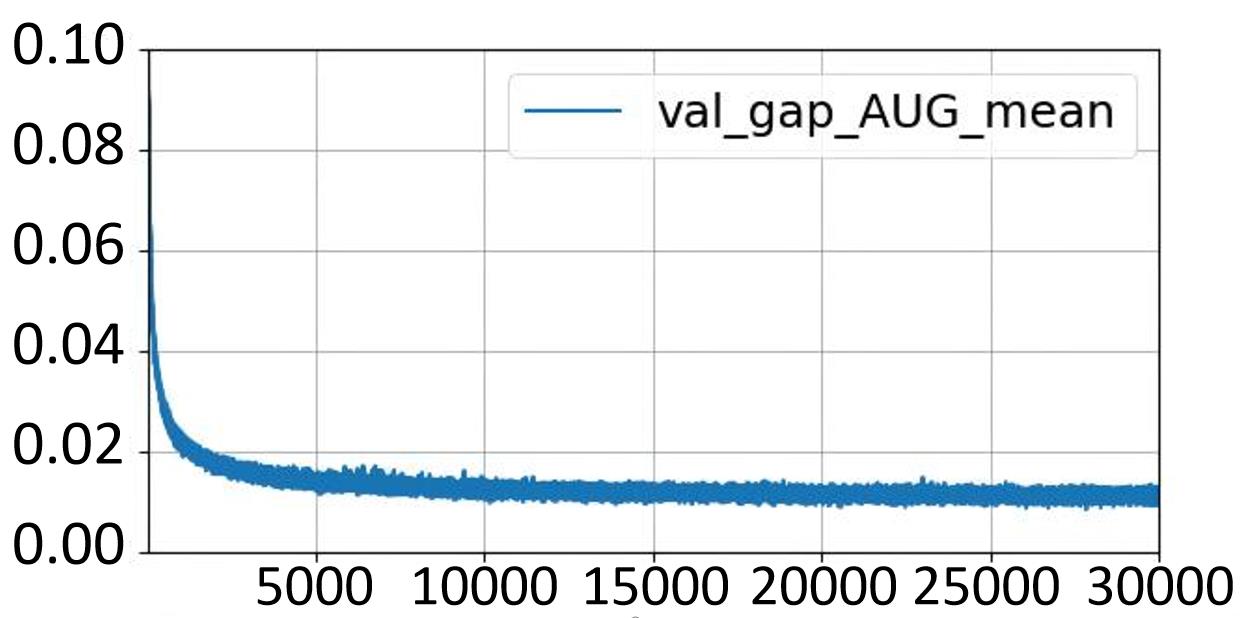}}
     \captionof{figure}{{Training curves of AMDKD-POMO for solving CVRP. The x-axis is the epoch and the y-axis is the average gaps on the used three exemplar distributions.}}
     \label{fig:convergence}
	\end{minipage}
	}
\end{table}

\subsection{Stability studies of AMDKD.}
\begin{figure}[H]
    \centering
    \includegraphics[width=\textwidth]{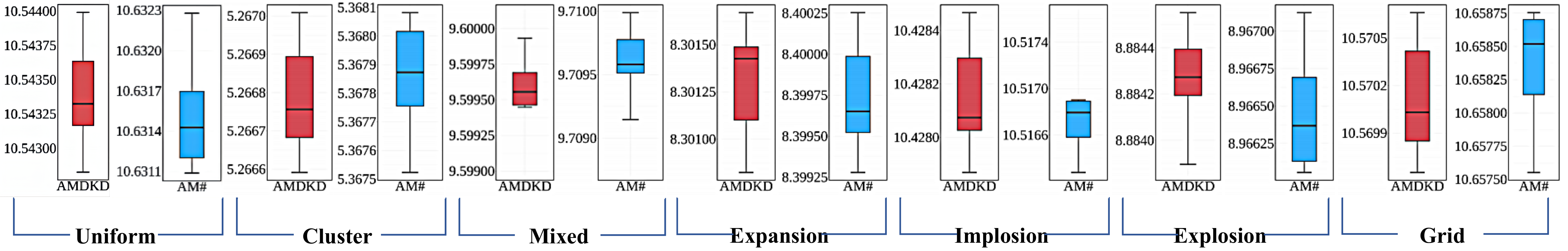}
    \vspace{-15pt}
    \caption{Experiment results of AMDKD-AM (red) and AM$^\#$ (blue) with different random seeds.}
    \label{fig:multi_test}
\end{figure}

To demonstrate the stability of our experiment results, we take CVRP-50 as an example and independently run our trained AMDKD-AM and AM$^\#$ for 10 times, where we adopt different random seeds during the sampling process. As shown in Figure \ref{fig:multi_test}, both AMDKD-AM and AM$^\#$ exhibit extremely small fluctuations (even less than 0.001) when running with different seeds. Based on the Wilcoxon test, our AMDKD-AM significantly (with $p$-value < 0.001) outperforms AM$^\#$ on every unseen in-distributions and OoD distributions. As for AMDKD-POMO and POMO$^\#$, the greedy decoding strategy is adopted, hence the results are expected to be stable with almost no fluctuation.

\section{Detailed results on benchmark datasets}
\label{app:lib}
We present the detailed results of benchmark dataset in Table \ref{tab:tsplib} (TSPLIB) and Table \ref{tab:cvrplib} (CVRPLIB), respectively.

As displayed, models that have only been trained on the uniform distribution, i.e., AM and POMO, perform extremely poorly when inferring instances that may follow unknown distributions from the benchmark. The upgraded POMO$^\#$ significantly outperforms POMO, which seemingly alleviates this issue through training on our exemplar distributions, however, this simple strategy does not work well with AM. Regarding the prior cross-distribution generalization methods including GANCO, HAC, PSRO and DROP which show the potential to improve AM or POMO, their results are still far from satisfactory, where the performance of DROP are even inferior to  POMO$^\#$ trained on our exemplar distributions. Nevertheless, our AMDKD not only outperforms all these baselines, but also brings a much more significant improvement over the backbone AM and POMO than those baselines do. 
{Meanwhile, our AMDKD-POMO exhibits a much better generalization over POMO$^\#$ on CVRPLIB.} 
Finally, we note that AMDKD equipped with EAS performs the best among all baselines, even achieving the optimal solution on some instances (e.g., KroA100, KroD100, lin105 in TSPLIB and X-n110-k14 in CVRPLIB). This leads to a new state-of-the-art performance for neural methods on these benchmark datasets. {Finally, we list the full generalization results of our AMDKD in {Table \ref{tab:fulltsplib}} (TSPLIB) and Table \ref{tab:fullcvrplib} (CVRPLIB), respectively.}

\begin{table}[htbp]
  \centering
  \vspace{-5pt}
    \caption{Detailed generalization results on selected instances from TSPLIB.}
    \vspace{5pt}
      \resizebox{0.99\textwidth}{!}{
    \begin{tabular}{c|c|c|ccccc|ccccc}
    \toprule
    \multicolumn{1}{c|}{Instance} & \multicolumn{1}{c|}{Opt.} & \multicolumn{1}{c|}{PSRO} & \multicolumn{1}{c}{AM} & \multicolumn{1}{c}{GANCO} & \multicolumn{1}{c}{HAC} & \multicolumn{1}{c}{AM$^\#$} & \multicolumn{1}{c|}{{AMDKD-AM}} & \multicolumn{1}{c}{POMO} & \multicolumn{1}{c}{DROP} & \multicolumn{1}{c}{{POMO$^\#$}} & \multicolumn{1}{c}{{AMDKD-POMO}} & \multicolumn{1}{c}{{AMDKD+EAS}} \\
    \midrule
    \multicolumn{1}{c|}{KroA100} & \multicolumn{1}{c|}{21282} & 21703  & \multicolumn{1}{c}{46621} & 21908  & 21838  & 22138  & 21650  & \multicolumn{1}{c}{38452} & \multicolumn{1}{c}{24623} & 21285 & 21285 & 21282 \\
    \multicolumn{1}{c|}{KroB100} & \multicolumn{1}{c|}{22141} & 22855  & \multicolumn{1}{c}{37921} & 22956  & 23110  & 23189  & 22350  & \multicolumn{1}{c}{33521} & \multicolumn{1}{c}{24874} & 22197 & 22233 & 22195 \\
    \multicolumn{1}{c|}{KroC100} & \multicolumn{1}{c|}{20749} & 21079  & \multicolumn{1}{c}{34258} & 21139  & 21068  & 22326  & 21279  & \multicolumn{1}{c}{30736} & \multicolumn{1}{c}{24785} & 20751 & 20752 & 20947 \\
    \multicolumn{1}{c|}{KroD100} & \multicolumn{1}{c|}{21294} & 21828  & \multicolumn{1}{c}{36141} & 21929  & 22625  & 23093  & 21863  & \multicolumn{1}{c}{29512} & \multicolumn{1}{c}{23257} & 21352 & 21314 & 21294 \\
    \multicolumn{1}{c|}{KroE100} & \multicolumn{1}{c|}{22068} & 22532  & \multicolumn{1}{c}{29628} & 23174  & 22807  & 22865  & 22327  & \multicolumn{1}{c}{26829} & \multicolumn{1}{c}{26057} & 22179 & 22185 & 22111 \\
    \multicolumn{1}{c|}{lin105} & \multicolumn{1}{c|}{14379} & 15372  & \multicolumn{1}{c}{15148} & 15478  & 15003  & 16865  & 14988  & \multicolumn{1}{c}{14922} & \multicolumn{1}{c}{14688} & 14430 & 14430 & 14379 \\
    \multicolumn{1}{c|}{pr107} & \multicolumn{1}{c|}{44303} & 45288  & \multicolumn{1}{c}{53846} & 45393  & 47250  & 76152  & 46146  & \multicolumn{1}{c}{52846} & \multicolumn{1}{c}{47853} & 44647 & 45022 & 44347 \\
    \midrule
    Avg. Gap & \multirow{2}{*}{0.00\%} & \multirow{2}{*}{2.93\%} & \multirow{2}{*}{55.19\%} & \multirow{2}{*}{3.80\%} & \multirow{2}{*}{4.16\%} & \multirow{2}{*}{16.80\%} & \multirow{2}{*}{{2.50\%}} & \multirow{2}{*}{37.63\%} & \multirow{2}{*}{12.14\%} & \multirow{2}{*}{0.31\%} & \multirow{2}{*}{{0.44\%}} & \multirow{2}{*}{{0.21\%}} \\
    ($n$=100-150) &&&&&&&&&&&&\\
    \midrule
    \multicolumn{1}{c|}{ch150} & \multicolumn{1}{c|}{6528} & 6866  & \multicolumn{1}{c}{6930} & 6704  & 6852  & 6680  & 6669  & \multicolumn{1}{c}{6844} & \multicolumn{1}{c}{6709} & 6574  & 6609  & 6554 \\
    \multicolumn{1}{c|}{rat195} & \multicolumn{1}{c|}{2323} & 2600  & \multicolumn{1}{c}{2612} & 2585  & 2638  & 3237  & 2550  & \multicolumn{1}{c}{2554} & \multicolumn{1}{c}{2403} & 2422  & 2432  & 2406 \\
    \multicolumn{1}{c|}{kroA200} & \multicolumn{1}{c|}{29368} & 31450  & \multicolumn{1}{c}{35637} & 31741  & 33174  & 34294  & 31112  & \multicolumn{1}{c}{34972} & \multicolumn{1}{c}{34275} & 29840 & 29906 & 29931 \\
    \midrule
    Avg. Gap & \multirow{2}{*}{0.00\%} &\multirow{2}{*}{ 8.06\%} & \multirow{2}{*}{13.32\%} & \multirow{2}{*}{7.36\%} & \multirow{2}{*}{10.50\%} & \multirow{2}{*}{19.48\%} & \multirow{2}{*}{{5.96\%}} & \multirow{2}{*}{11.29\%} & \multirow{2}{*}{7.64\%} & \multirow{2}{*}{2.19\%} & \multirow{2}{*}{2.59\%} & \multirow{2}{*}{1.96\%} \\
    ($n$=150-200) &&&&&&&&&&&&\\
    \bottomrule
    \end{tabular}%
    }
  \label{tab:tsplib}%
\end{table}%

\begin{table}[htbp]
  \centering
   \setlength\tabcolsep{10pt}
   \vspace{-5pt}
  \caption{Detailed generalization results on selected instances from CVRPLIB.}
   \vspace{5pt}
  \resizebox{0.99\textwidth}{!}{
    \begin{tabular}{c|c|ccc|ccccc}
    \toprule
    \multicolumn{1}{c|}{Instance} & \multicolumn{1}{c|}{Opt.} & \multicolumn{1}{c}{AM} & \multicolumn{1}{c}{AM$^\#$} & \multicolumn{1}{c|}{AMDKD-AM} & \multicolumn{1}{c}{POMO} & \multicolumn{1}{c}{DROP} & \multicolumn{1}{c}{POMO$^\#$} & \multicolumn{1}{c}{AMDKD-POMO} & \multicolumn{1}{c}{AMDKD+EAS} \\
    \midrule
    X-n101-k25 & 27591 & 38264 & 30327 & 30782 & 29484 & 28949 & 30510 & 29299 & 27855 \\
    X-n106-k14 & 26362 & 27923 & 27958 & 27279 & 27762 & 27308 & 27077 & 26847 & 26550 \\
    X-n110-k13 & 14971 & 16320 & 15668 & 15348 & 15896 & 15386 & 15175 & 15315 & 14971 \\
    X-n115-k10 & 12747 & 14055 & 14638 & 13366 & 13952 & 13783 & 13609 & 13418 & 12883 \\
    X-n120-k6 & 13332 & 14456 & 16094 & 14162 & 14351 & 14058 & 13997 & 13604 & 13457 \\
    X-n125-k30 & 55539 & 74329 & 68870 & 58507 & 69560 & 61382 & 62383 & 58570 & 56596 \\
    X-n129-k18 & 28940 & 30869 & 30833 & 29851 & 30155 & 30075 & 29597 & 29449 & 29007 \\
    X-n134-k13 & 10916 & 13952 & 12709 & 12573 & 13483 & 12846 & 11325 & 11330 & 11073 \\
    X-n139-k10 & 13590 & 14893 & 14953 & 14097 & 14132 & 13979 & 14053 & 13955 & 13704 \\
    X-n143-k7 & 15700 & 18251 & 18345 & 16509 & 17923 & 17682 & 16487 & 16346 & 15871 \\
    \midrule
    Avg. Gap & \multirow{2}{*}{0.00\%} & \multirow{2}{*}{16.65\%} & \multirow{2}{*}{13.00\%} & \multirow{2}{*}{{6.12\%}} & \multirow{2}{*}{10.66\%} & \multirow{2}{*}{7.25\%} & \multirow{2}{*}{5.32\%} & \multirow{2}{*}{{3.54\%}} & \multirow{2}{*}{{0.92\%}} \\
    ($n=$100-150) &&&&&&&&&\\
    \midrule
    X-n153-k22 & 21220 & 38423 & 24722 & 23766 & 26386 & 24386 & 23629 & 23590 & 21849 \\
    X-n157-k13 & 16876 & 22051 & 19890 & 17539 & 19978 & 18378 & 17950 & 17450 & 17093 \\
    X-n181-k23 & 25569 & 27826 & 27314 & 26415 & 27428 & 27094 & 29014 & 26756 & 25736 \\
    X-n190-k8 & 16980 & 37820 & 21020 & 21162 & 22310 & 19864 & 18912 & 17575 & 17228 \\
    X-n200-k36 & 58578 & 76528 & 66298 & 62335 & 73135 & 64921 & 62228 & 62967 & 60562 \\
    \midrule
    Avg. Gap & \multirow{2}{*}{0.00\%} & \multirow{2}{*}{54.79\%} & \multirow{2}{*}{15.63\%} & \multirow{2}{*}{{10.06\%}} & \multirow{2}{*}{21.25\%} & \multirow{2}{*}{11.52\%} & \multirow{2}{*}{9.76\%} & \multirow{2}{*}{{6.04\%}} & \multirow{2}{*}{{1.95\%}} \\
    ($n=$150-200) &&&&&&&&&\\
    \bottomrule
    \end{tabular}%
    }
  \label{tab:cvrplib}%
\end{table}%

\begin{table}[htbp]
  \centering
  \caption{{Full generalization results on TSPLIB (instances ranged from 100 to 200).}}
     \setlength\tabcolsep{8pt}
   \vspace{5pt}
  \resizebox{0.99\textwidth}{!}{
    \begin{tabular}{cc|cccc|cccccc}
    \toprule
    \multicolumn{1}{c|}{\multirow{2}[2]{*}{Instance}} & \multirow{2}[2]{*}{Opt.} & \multicolumn{2}{c}{AM$^\#$} & \multicolumn{2}{c|}{AMDKD-AM} & \multicolumn{2}{c}{{POMO$^\#$}} & \multicolumn{2}{c}{{AMDKD-POMO}} & \multicolumn{2}{c}{AMDKD+EAS} \\
    \multicolumn{1}{c|}{} &       & Obj.  & Gap   & Obj.  & Gap   & Obj.  & Gap   & Obj.  & Gap   & Obj.  & Gap \\
    \midrule
    \multicolumn{1}{c|}{kroA100} & 21282 & 22138  & 4.02\% & 21650  & 1.73\% & 21285 & 0.02\% & 21285 & 0.02\% & 21282 & 0.00\% \\
    \multicolumn{1}{c|}{kroB100} & 22141 & 23189  & 4.73\% & 22350  & 0.94\% & 22197 & 0.25\% & 22233 & 0.41\% & 22195 & 0.24\% \\
    \multicolumn{1}{c|}{kroC100} & 20749 & 22326  & 7.60\% & 21279  & 2.55\% & 20751 & 0.01\% & 20752 & 0.02\% & 20947 & 0.95\% \\
    \multicolumn{1}{c|}{kroD100} & 21294 & 23093  & 8.45\% & 21863  & 2.67\% & 21352 & 0.27\% & 21314 & 0.09\% & 21294 & 0.00\% \\
    \multicolumn{1}{c|}{kroE100} & 22068 & 22865  & 3.61\% & 22327  & 1.17\% & 22179 & 0.50\% & 22185 & 0.53\% & 22111 & 0.19\% \\
    \multicolumn{1}{c|}{eil101} & 629   & 663   & 5.45\% & 647   & 2.82\% & 641   & 1.90\% & 645   & 2.55\% & 629   & 0.00\% \\
    \multicolumn{1}{c|}{lin105} & 14379 & 16865  & 17.29\% & 14988  & 4.24\% & 14430 & 0.36\% & 14430 & 0.36\% & 14379 & 0.00\% \\
    \multicolumn{1}{c|}{pr107} & 44303 & 76152  & 71.89\% & 46146  & 4.16\% & 44647 & 0.78\% & 45022 & 1.62\% & 44347 & 0.10\% \\
    \multicolumn{1}{c|}{pr124} & 59030 & 62075  & 5.16\% & 60042  & 1.71\% & 59031 & 0.00\% & 59281 & 0.43\% & 59030 & 0.00\% \\
    \multicolumn{1}{c|}{bier127} & 118282 & 275748  & 133.13\% & 123211  & 4.17\% & 119232 & 0.80\% & 119052 & 0.65\% & 118729 & 0.38\% \\
    \multicolumn{1}{c|}{ch130} & 6110  & 6231  & 1.98\% & 6171  & 1.00\% & 6146  & 0.60\% & 6152  & 0.69\% & 6115  & 0.08\% \\
    \multicolumn{1}{c|}{pr136} & 96772 & 100194  & 3.54\% & 99912  & 3.24\% & 98478 & 1.76\% & 98215 & 1.49\% & 97487 & 0.74\% \\
    \multicolumn{1}{c|}{pr144} & 58537 & 66628  & 13.82\% & 60807  & 3.88\% & 59034 & 0.85\% & 58956 & 0.72\% & 58794 & 0.44\% \\
    \multicolumn{1}{c|}{ch150} & 6528  & 6680  & 2.33\% & 6669  & 2.16\% & 6574  & 0.70\% & 6609  & 1.25\% & 6554  & 0.40\% \\
    \multicolumn{1}{c|}{kroA150} & 26524 & 29501  & 11.22\% & 27354  & 3.13\% & 26723 & 0.75\% & 26808 & 1.07\% & 26538 & 0.05\% \\
    \multicolumn{1}{c|}{kroB150} & 26130 & 28585  & 9.39\% & 26820  & 2.64\% & 26334 & 0.78\% & 26328 & 0.76\% & 26152 & 0.08\% \\
    \multicolumn{1}{c|}{pr152} & 73682 & 85703  & 16.31\% & 78120  & 6.02\% & 74673 & 1.35\% & 75270 & 2.16\% & 75250 & 2.13\% \\
    \multicolumn{1}{c|}{rat195} & 2323  & 3237  & 39.35\% & 2550  & 9.77\% & 2422  & 4.25\% & 2432  & 4.68\% & 2406  & 3.57\% \\
    \multicolumn{1}{c|}{kroA200} & 29368 & 34294  & 16.77\% & 31112  & 5.94\% & 29840 & 1.61\% & 29906 & 1.83\% & 29931 & 1.92\% \\
    \multicolumn{1}{c|}{kroB200} & 29437 & 34074  & 15.75\% & 31968  & 8.60\% & 29665 & 0.77\% & 30132 & 2.36\% & 29765 & 1.11\% \\
    \midrule
    \multicolumn{2}{c|}{Avg. Gap} & \multicolumn{2}{c}{19.59\%} & \multicolumn{2}{c|}{3.63\%} & \multicolumn{2}{c}{0.92\%} & \multicolumn{2}{c}{1.18\%} & \multicolumn{2}{c}{0.62\%} \\

    \bottomrule
    \end{tabular}%
    }
  \label{tab:fulltsplib}%
\end{table}%

\begin{table}[htbp]
  \centering
  \caption{{Full generalization results on CVRPLIB (instances ranged from 100 to 200).}}
     \setlength\tabcolsep{8pt}
   \vspace{5pt}
  \resizebox{0.99\textwidth}{!}{
    \begin{tabular}{cc|cccc|cccccc}
    \toprule
    \multicolumn{1}{c|}{\multirow{2}[2]{*}{Instance}} & \multirow{2}[2]{*}{Opt.} & \multicolumn{2}{c}{AM$^\#$} & \multicolumn{2}{c|}{AMDKD-AM} & \multicolumn{2}{c}{POMO$^\#$} & \multicolumn{2}{c}{AMDKD-POMO} & \multicolumn{2}{c}{AMDKD+EAS} \\
    \multicolumn{1}{c|}{} &       & Obj.  & Gap   & Obj.  & Gap   & Obj.  & Gap   & Obj.  & Gap   & Obj.  & Gap \\
    \midrule
    \multicolumn{1}{c|}{X-n101-k25} & 27591 & 30327 & 9.92\% & \multicolumn{1}{r}{30782 } & \multicolumn{1}{r|}{11.57\%} & 30510  & 10.58\% & 29299  & 6.19\% & 27855 & 0.96\% \\
    \multicolumn{1}{c|}{X-n106-k14} & 26362 & 27958 & 6.06\% & \multicolumn{1}{r}{27279 } & \multicolumn{1}{r|}{3.48\%} & 27077  & 2.71\% & 26847  & 1.84\% & 26550 & 0.71\% \\
    \multicolumn{1}{c|}{X-n110-k13} & 14971 & 15668 & 4.66\% & \multicolumn{1}{r}{15348 } & \multicolumn{1}{r|}{2.52\%} & 15175  & 1.36\% & 15315  & 2.30\% & 14971 & 0.00\% \\
    \multicolumn{1}{c|}{X-n115-k10} & 12747 & 14638 & 14.83\% & \multicolumn{1}{r}{13366 } & \multicolumn{1}{r|}{4.86\%} & 13609  & 6.76\% & 13418  & 5.27\% & 12883 & 1.07\% \\
    \multicolumn{1}{c|}{X-n120-k6} & 13332 & 16094 & 20.71\% & \multicolumn{1}{r}{14162 } & \multicolumn{1}{r|}{6.23\%} & 13997  & 4.99\% & 13604  & 2.04\% & 13457 & 0.94\% \\
    \multicolumn{1}{c|}{X-n125-k30} & 55539 & 68870 & 24.00\% & \multicolumn{1}{r}{58507 } & \multicolumn{1}{r|}{5.34\%} & 62383  & 12.32\% & 58570  & 5.46\% & 56596 & 1.90\% \\
    \multicolumn{1}{c|}{X-n129-k18} & 28940 & 30833 & 6.54\% & \multicolumn{1}{r}{29851 } & \multicolumn{1}{r|}{3.15\%} & 29597  & 2.27\% & 29449  & 1.76\% & 29007 & 0.23\% \\
    \multicolumn{1}{c|}{X-n134-k13} & 10916 & 12709 & 16.43\% & \multicolumn{1}{r}{12573 } & \multicolumn{1}{r|}{15.18\%} & 11325  & 3.74\% & 11330  & 3.79\% & 11073 & 1.44\% \\
    \multicolumn{1}{c|}{X-n139-k10} & 13590 & 14953 & 10.03\% & \multicolumn{1}{r}{14097 } & \multicolumn{1}{r|}{3.73\%} & 14053  & 3.41\% & 13955  & 2.69\% & 13704 & 0.84\% \\
    \multicolumn{1}{c|}{X-n143-k7} & 15700 & 18345 & 16.84\% & \multicolumn{1}{r}{16509 } & \multicolumn{1}{r|}{5.15\%} & 16487  & 5.01\% & 16346  & 4.12\% & 15871 & 1.09\% \\
    \multicolumn{1}{c|}{X-n148-k46} & 43448 & 61800 & 42.24\% & \multicolumn{1}{r}{52627 } & \multicolumn{1}{r|}{21.13\%} & 53217  & 22.48\% & 46993  & 8.16\% & 44075 & 1.44\% \\
    \multicolumn{1}{c|}{X-n153-k22} & 21220 & 24722 & 16.50\% & \multicolumn{1}{r}{23766 } & \multicolumn{1}{r|}{12.00\%} & 23629  & 11.35\% & 23590  & 11.17\% & 21849 & 2.96\% \\
    \multicolumn{1}{c|}{X-n157-k13} & 16876 & 19890 & 17.86\% & \multicolumn{1}{r}{17539 } & \multicolumn{1}{r|}{3.93\%} & 17950  & 6.36\% & 17450  & 3.40\% & 17093 & 1.29\% \\
    \multicolumn{1}{c|}{X-n162-k11} & 14138 & 14762 & 4.41\% & \multicolumn{1}{r}{14663 } & \multicolumn{1}{r|}{3.72\%} & 14951  & 5.75\% & 14903  & 5.41\% & 14543 & 2.86\% \\
    \multicolumn{1}{c|}{X-n167-k10} & 20557 & 21686 & 5.49\% & \multicolumn{1}{r}{21468 } & \multicolumn{1}{r|}{4.43\%} & 21573  & 4.94\% & 21401  & 4.11\% & 20890 & 1.62\% \\
    \multicolumn{1}{c|}{X-n172-k51} & 45607 & 62419 & 36.86\% & \multicolumn{1}{r}{64444 } & \multicolumn{1}{r|}{41.30\%} & 49844  & 9.29\% & 49741  & 9.06\% & 46340 & 1.61\% \\
    \multicolumn{1}{c|}{X-n176-k26} & 47812 & 53263 & 11.40\% & \multicolumn{1}{r}{51102 } & \multicolumn{1}{r|}{6.88\%} & 54149  & 13.25\% & 53189  & 11.25\% & 49241 & 2.99\% \\
    \multicolumn{1}{c|}{X-n181-k23} & 25569 & 27314 & 6.83\% & \multicolumn{1}{r}{26415 } & \multicolumn{1}{r|}{3.31\%} & 29014  & 13.47\% & 26756  & 4.64\% & 25736 & 0.65\% \\
    \multicolumn{1}{c|}{X-n186-k15} & 24145 & 25845 & 7.04\% & \multicolumn{1}{r}{25526 } & \multicolumn{1}{r|}{5.72\%} & 25827  & 6.97\% & 26332  & 9.06\% & 24893 & 3.10\% \\
    \multicolumn{1}{c|}{X-n190-k8} & 16980 & 21020 & 23.79\% & \multicolumn{1}{r}{21162 } & \multicolumn{1}{r|}{24.63\%} & 18912  & 11.38\% & 17575  & 3.50\% & 17228 & 1.46\% \\
    \multicolumn{1}{c|}{X-n195-k51} & 44225 & 57830 & 30.76\% & \multicolumn{1}{r}{60882 } & \multicolumn{1}{r|}{37.66\%} & 48907  & 10.59\% & 51284  & 15.96\% & 45758 & 3.47\% \\
    \midrule
    \multicolumn{2}{c|}{Avg. Gap} & \multicolumn{2}{c}{15.87\%} & \multicolumn{2}{c|}{{10.76\%}} & \multicolumn{2}{c}{8.05\%} & \multicolumn{2}{c}{{5.77\%}} & \multicolumn{2}{c}{{1.55\%}} \\
    \bottomrule
    \end{tabular}%
    }
  \label{tab:fullcvrplib}%
\end{table}%

\section{Used assets and licenses}
\label{app:assert}

Table \ref{tab:asset} lists the used assets in our work, which are all open-source for academic research. For our code and used data (new assets), we are using the MIT License.

\begin{table}[htbp]
  \centering
  \setlength\tabcolsep{9pt}
  \caption{Used assets and their licenses.}
  \vspace{4pt}
  \begin{threeparttable}
    \resizebox{0.8\textwidth}{!}{
    \begin{tabular}{cccc}
    \toprule
    Type  & Asset & License & Usage \\
    \midrule
    \multirow{9}[2]{*}{Code} & Gurobi \cite{gurobi} & Free Academic lisence & Evaluation \\
          & LKH3 \cite{lkh3}  & Available for academic use & Evaluation \\
          & AM \cite{kool2018attention}    & MIT License & Remodification and evaluation \\
          & POMO \cite{kwon2020pomo}  & MIT License & Remodification and evaluation \\
          & LCP \cite{Kim2021LearningCP}   & MIT License & Remodification and evaluation \\
          & HAC \cite{Zhang2022LearningTS}   & MIT License & Remodification and evaluation \\
          & EAS \cite{eas}   & MIT License & Remodification and evaluation \\
          & DACT \cite{ma2021learning}  & MIT License & Remodification and evaluation \\
          & tspgen \cite{Bossek2019EvolvingDT} & GNU General Public License v3.0 & Generating datasets \\
    \midrule
    \multirow{2}[2]{*}{Datasets} & TSPLIB \cite{reinelt1991tsplib} & Available for any non-commerial use & Testing \\
          & CVRPLIB \cite{uchoa2017new} & Available for any non-commerial use & Testing \\
    \bottomrule
    \end{tabular}%
    }
    \end{threeparttable}
  \label{tab:asset}%
\end{table}%

\end{document}